\documentclass[10pt,journal,compsoc]{IEEEtran}
\ifCLASSOPTIONcompsoc
  \usepackage[nocompress]{cite}
\else
  \usepackage{cite}
\fi
\ifCLASSINFOpdf
  \usepackage[pdftex]{graphicx}
  \graphicspath{{Figures_and_Plots/}}
  \DeclareGraphicsExtensions{.pdf,.jpg,.png}
\else
\fi
\usepackage{amsmath}
\usepackage{amssymb}

\usepackage{mleftright}
\mleftright

\usepackage[noend]{algorithmic}

\usepackage{array}

\ifCLASSOPTIONcompsoc
  \usepackage[caption=false,font=footnotesize,labelfont=sf,textfont=sf]{subfig}
\else
  \usepackage[caption=false,font=footnotesize]{subfig}
\fi
\usepackage{url}

\usepackage[utf8]{inputenc} 
\usepackage{pgfplots}
\DeclareUnicodeCharacter{2212}{−}
\usepgfplotslibrary{groupplots,dateplot}
\usetikzlibrary{patterns,shapes.arrows}
\pgfplotsset{compat=newest}

\usepackage{multirow} 

\usepackage{tikz}

\begin{document}
%
\title{Volumetric Calculation of Quantization Error in 3-D Vision Systems}
%
%
%
%

\author{Eleni~Bohacek,
        Andrew~J.~Coates,
        and~David~R.~Selviah
\IEEEcompsocitemizethanks{\IEEEcompsocthanksitem E. Bohacek and D. R. Selviah are with the Department of Electronic and Electrical Engineering, University College London (UCL), WC1E 7JE, United Kingdom.\protect\\
E-mail: eleni.bohacek.10@ucl.ac.uk
\IEEEcompsocthanksitem A. J. Coates is with the Mullard Space Science Laboratory, University College London.}
}
\IEEEtitleabstractindextext{%
\begin{abstract}
This paper investigates how the inherent quantization of camera sensors introduces uncertainty in the calculated position of an observed feature during 3-D mapping. It is typically assumed that pixels and scene features are points, however, a pixel is a two-dimensional area that maps onto multiple points in the scene. This uncertainty region is a bound for quantization error in the calculated point positions. Earlier studies calculated the volume of two intersecting pixel views, approximated as a cuboid, by projecting pyramids and cones from the pixels into the scene. In this paper, we reverse this approach by generating an array of scene points and calculating which scene points are detected by which pixel in each camera. This enables us to map the uncertainty regions for every pixel correspondence for a given camera system in one calculation, without approximating the complex shapes. The dependence of the volumes of the uncertainty regions on camera baseline length, focal length, pixel size, and distance to object, shows that earlier studies overestimated the quantization error by at least a factor of two. For static camera systems the method can also be used to determine volumetric scene geometry without the need to calculate disparity maps.
\end{abstract}

}

\maketitle

\IEEEdisplaynontitleabstractindextext

%
\IEEEpeerreviewmaketitle

\IEEEraisesectionheading{\section{Introduction}\label{sec:introduction}}

%
%
%
%
%
\IEEEPARstart{T}{hree} dimensional (3-D) scene reconstruction from multiple views is one of the oldest fields of computer vision and photogrammetry, and it has diverse applications including autonomous vehicle navigation, historical conservation, and industrial tasks such as process inspection and quality assurance \cite{van_damme_computer_2015,yastikli_documentation_2007,khoshelham_accuracy_2012}. Stereo reconstruction is a process which calculates the position of an object feature in 3-D space by detecting it in two images. This can be summarized in the following steps:
\begin{enumerate}
\item Geometric calibration determines the intrinsic and extrinsic camera parameters.
\item Correspondence matching identifies the pairs of pixels that view the same object.
\item Triangulation calculates the 3-D position of the object using disparity in the image plane.
\end{enumerate}

Point clouds (and depth or disparity maps) are the most primitive 3-D data calculated from these processes. Every point in a point cloud has an x, y, z value in Euclidean space and a brightness (color or black and white). However, this 3-D information is meaningless if we have not characterized its associated uncertainty, reliability, reproducibility, and errors. The uncertainty in brightness is determined by radiometric factors such as the surface properties and illumination conditions \cite{sonka_image_2014,kamberova_sensor_1998}. The uncertainty in position is caused by errors at all three stages of stereo reconstruction, system geometry and image plane quantization.

A lot of research has been carried out into the study of errors generated from calibration \cite{xu_error_2013, schreve_how_2014, di_leo_uncertainty_2011, yang_analysis_2017, shawash_real-time_2013} and correspondence matching \cite{ng_solving_2019, liu_3d_2015, ma_robust_2014, scharstein_taxonomy_2001, li_comparison_2011, marr_cooperative_1976}. There is also a disparate body of research that we summarize as geometrical constraints on error, which is concerned with the fundamental limitations on what a given system of cameras can measure. These factors include intrinsic camera properties, such as the sensor dimensions and focal length, and extrinsic camera properties, such as their baseline distance and angle of focal axis convergence \cite{kyto_method_2011, schreve_how_2014, yang_analysis_2017}. Another geometrical constraint is the quantization of the image plane into pixels by the camera sensor (such as a CCD or CMOS), which gives rise to errors in point position called quantization error \cite{fooladgar_geometrical_2013, wu_analysis_1998}.

\subsection{Image Plane Quantization Error}
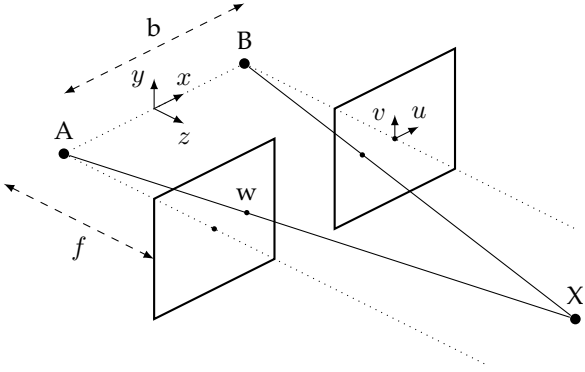
\begin{figure}[!t]
\centering
     \scalebox{1}{\begin{tikzpicture}[y={(0cm,0.8cm)},x={(-0.8cm,-0.4cm)}, z={(0.8cm,-0.4cm)}]

\coordinate (O) at (0, 0, 0);
\draw[-latex] (O) -- +(-0.5, 0,  0) node [above] {$x$};
\draw[-latex] (O) -- +(0, 0.5, 0) node [left] {$y$};
\draw[-latex] (O) -- +(0, 0, 0.5) node [below] {$z$};
\node [circle,fill=black,inner sep=0pt,minimum size=4pt,label=north:{B}] (camA) at (-1.5,0,0) {}; 
\node [circle,fill=black,inner sep=0pt,minimum size=4pt,label=north:{A}] (camB) at (1.5,0,0) {}; 
\draw [dotted] (camA) -- (camB);
\draw [thick] (0.5,-1,2.5) -- (0.5,1,2.5) -- (2.5,1,2.5) -- (2.5,-1,2.5) -- cycle;
\draw [thick] (-0.5,-1,2.5) -- (-0.5,1,2.5) -- (-2.5,1,2.5) -- (-2.5,-1,2.5) -- cycle;
\node [circle,fill=black,inner sep=0pt,minimum size=2pt] (v1) at (-1.5,0,2.5) {}; 
\node [circle,fill=black,inner sep=0pt,minimum size=2pt] (v1) at (1.5,0,2.5) {}; 
\draw [dotted] (camA) -- (-1.5,0,5.5);
\draw [dotted] (camB) -- (1.5,0,7);
\coordinate (O) at (-1.5, 0, 2.5);
\draw[-latex] (O) -- +(-0.4, 0,  0) node [above] {$u$};
\draw[-latex] (O) -- +(0, 0.4, 0) node [left] {$v$};
\node [circle,fill=black,inner sep=0pt,minimum size=4pt,label=north:{X}] (pointX) at (0,0,7) {};
\draw [thin] (camA) -- (pointX);
\draw [thin] (camB) -- (pointX);

\node [circle,fill=black,inner sep=0pt,minimum size=2pt] (v1) at (-0.96,0,2.5) {}; 
\node [circle,fill=black,inner sep=0pt,minimum size=2pt,label=north:{w}] (v1) at (0.96,0,2.5) {}; 

\draw [latex-latex, dashed] (1.5,1,0) -- (-1.5,1,0);
\node [label=north:{b}] (baseline) at (0,0.9,0) {};
\draw [latex-latex, dashed] (2.5,0,0) -- (2.5,0,2.5);
\node [label=north:{$f$}] (baseline) at (2.5,-0.95,1.25) {};
\end{tikzpicture}}  

\caption{Perspective projection geometry for two coplanar cameras.}
\label{fig_perspective}
\end{figure}

Many 3-D from stereo algorithms use the disparity in the image planes to calculate 3-D position using triangulation \cite{hartley_triangulation_1997}. This assumes that two lines from the perspective center of each camera, $A$ and $B$ in figure \ref{fig_perspective}, intersect at the object position in space, denoted by $X$. These lines intersect with the image planes at $w_{A}$ and $w_{B}$, this is further discussed in section \ref{sec:pinhole}. By this stage of the calculation, errors in the object position on the image plane have been introduced, which may produce projection lines that do not intersect \cite{wenhardt_minimizing_2007}, motivating studies of how to minimize and adjust the errors in the image plane \cite{zhang_triangulation_2009, hartley_triangulation_1997}. Furthermore, the quantization of the camera sensor introduces limitations in the spatial resolution of the image plane, which is problematic because triangulation assumes pixels to be points rather than rectangles \cite{wu_analysis_1998}. 


\begin{figure}[!t]
\centering
\includegraphics[scale=0.9,trim=0 40 70 0, clip]{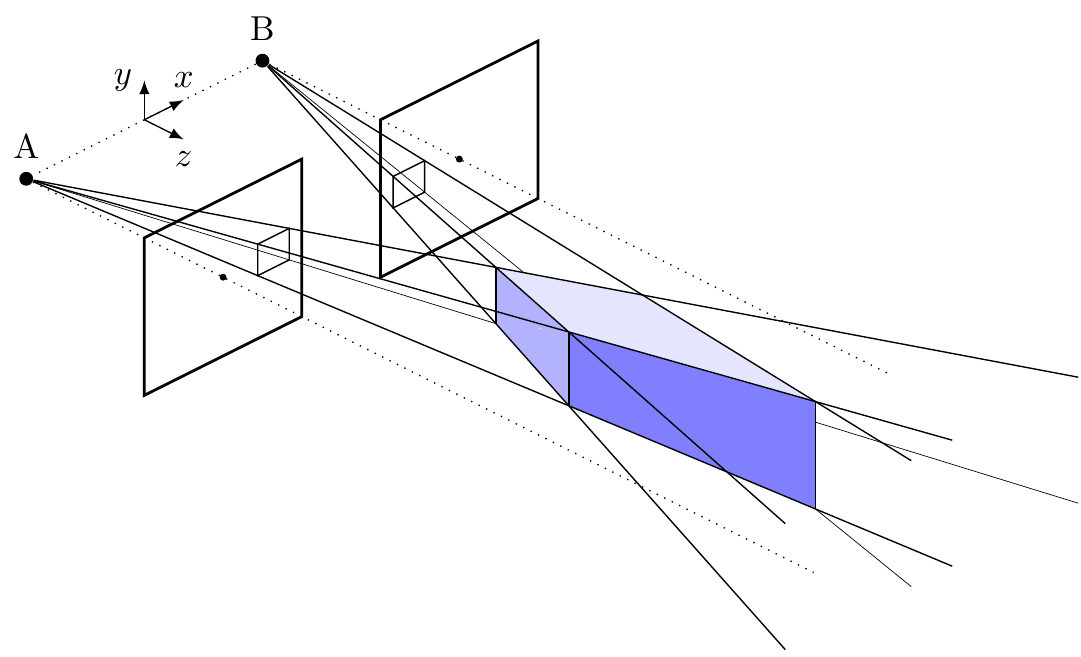}
\caption{Perspective projection of pyramids to represent pixel views. The intersection of these pyramids is a polyhedron.}
\label{fig_pyramids}
\end{figure}

If we model a pixel as a rectangle instead of a point, and its projection into space as a pyramid instead of a ray during triangulation, we can calculate the localization error caused by image plane quantization. The intersection of two pyramids gives a polyhedron of six or more sides, which represents an absolute uncertainty bound for the calculated object position, illustrated in fig. \ref{fig_pyramids}. This is referred to as an uncertainty region or error volume \cite{fooladgar_geometrical_2013,wu_analysis_1998}.

Wu, Sharma and Huang pioneered this approach in 1998, and used the volume of the polyhedron as an error measurement \cite{wu_analysis_1998}. Limited by computational power at the time, rather than calculating the volume of the polyhedron, they fit the vertices to an ellipsoid using principle component analysis and worked with the ellipsoid volume instead. They also suggested that the ratio of major to minor axis and the direction and size of the major axis gives a quantitative measure of localization error due to quantization.

The idea was picked up again by Fooladgar et al. in 2013, where they applied the idea to circular pixels, which makes the field of view a cone instead of a pyramid \cite{fooladgar_geometrical_2013}. The intersection of two cones is a non-trivial problem that can be solved using a Lagrangian method. They further proposed two analytic solutions that reduce the computational cost by assuming one pixel projection is a cone and the other is four planes, the faces of a pyramid. Due to computational limitations they took the took the maximum and minimum coordinate values of the polyhedron vertices to approximate this shape as a cuboid. Both the polyhedron volume and the cuboid volume have the same dimensions in $x$, $y$ and $z$, as shown in fig. \ref{fig_bounds}. The volume of the polyhedron (green) will always be less than the cuboid (blue), depending on the nature of the shape this can be a large or a small difference. The ellipsoid volume used in \cite{wu_analysis_1998} would be fitted to the vertices of the polyhedron. 

Both studies present methods to calculate ellipsoid or cuboid approximations of the polyhedral error volume, for one pixel combination at a time. In this paper, we present a method to calculate the true polyhedron volume for all pixel combinations in a stereo pair. Rather than calculating the complex polyhedral shape of each intersection by projecting pyramids or cones, our method generates a three-dimensional array of gridpoints in the scene and determines which gridpoints are within each intersection. This is the inverse problem of the previous approaches and the discretization of scene space is a technique used in occupancy grid mapping (see section \ref{sec:occupancy}) and volumetric scene reconstruction \ref{sec_Visual_Hull} . Given each camera we calculate which gridpoints are seen by each pixel and perform an intersection of these sets to obtain every uncertainty volume within that space. The method can be used for $n$ cameras with any rotation or translation between them (provided that their views overlap), but for this study we demonstrate the method using two coplanar camera views. In line with the earlier studies we investigate the effects of camera separation (baseline distance), focal length, pixel size and gridpoint position on the polyhedral and cuboid uncertainty volumes, and compared the latter results with \cite{fooladgar_geometrical_2013}. 

\subsection{Related Work}
\subsubsection{Geometric Constraints on Localization Error}

Modelling and predicting the accuracy of 3-D from stereo calculations has been a common goal of many studies over the last 40 years. Here we explore a subset of that research, concerning “geometrical constraints on error” as defined in section \ref{sec:introduction}.

One study presents an empirically derived model for the total absolute error in the $x$, $y$, and $z$ coordinates (as defined in fig. \ref{fig_perspective}) of a measured point, whereby the error for each coordinate is a sum of systematic and random components \cite{sankowski_estimation_2017}. They predict a similar total absolute error for the $x$ and $y$ coordinates and an order of magnitude greater error for the $z$ coordinate, although by their own admission these estimates are "pessimistically high". Even if not using this method, stating the error in each Cartesian direction as a confidence interval (for example $z = 2 \pm 0.1 \, m$) is a typical approach.

An equation of the following form has been used to define the total error of a point $\varepsilon$,

\begin{equation}\label{eq:error}
\varepsilon = \sqrt{\varepsilon_{x}^2 + \varepsilon_{y}^2 + \varepsilon_{z}^2}
\end{equation}

whereby $\varepsilon_{x}$ represents the variance of the $x$ coordinate \cite{schreve_how_2014} or $\varepsilon_{x}$ represents simulated absolute errors of the $x$ coordinate \cite{xu_error_2013, yang_analysis_2017, rodriguez-quinonez_improve_2017}. Physically this represents the magnitude of the vector connecting the true and measured location of the point. The magnitude of the three components can be very different (see section \ref{sec:discussion}) and this dimensionality information is lost when calculating this quantity, as this effectively this models the uncertainty volume as a sphere.

A model by Yang et al. shows how baseline distance, focal length, and convergence angle separately affect $\varepsilon$, and also uses a transfer function to see how they work in combination \cite{yang_analysis_2017}. Kyt\"o et al. performed an experiment with a multi-level calibration target to examine the effect of baseline distance and focal length on just depth error $\varepsilon_{z}$ \cite{kyto_method_2011}. Similarly, Ortiz et al. only looked at depth error but varied the resolution setting of their camera \cite{ortiz_depth_2018}.

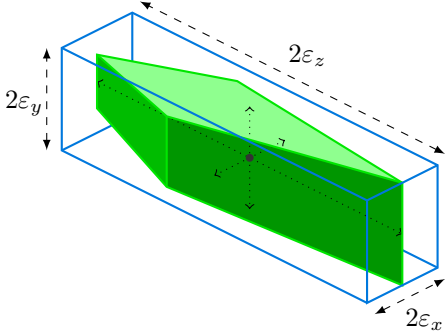
\begin{figure}[!t]
\centering
     \scalebox{1}{\begin{tikzpicture}[y={(0cm,1cm)},x={(-1cm,-0.5cm)}, z={(1cm,-0.5cm)}]

\definecolor{color0}{rgb}{0,0.469313725490196,0.8667}

\draw[thick,color0](0.46875, -0.681818,4.46429) -- (-0.46875, -0.681818,4.46429) -- (-0.46875, 0.681818,4.46429);
\draw[thick,color0](-0.46875, -0.681818,4.46429) -- (-0.46875, -0.681818,8.52273);

\fill [green!43.853137254902](0,0.357143,4.46429) -- (0.46875,-0.46875,5.85938) -- (0,0.681818,8.52273) -- (-0.46875,0.46875,5.85938) -- cycle;
\fill [green!73.853137254902!black](0,0.357143,4.46429) -- (0,-0.357143,4.46429) -- (0.46875,-0.46875,5.85938) -- (0.46875,0.46875,5.85938) -- cycle;
\fill [green!58.853137254902!black](0.46875,-0.46875,5.85938) -- (0.46875,0.46875,5.85938) -- (0,0.681818,8.52273) -- (0,-0.681818,8.52273) -- cycle;

\draw[thin,dotted, <->](-0.46875, 0,6.49351) -- (0.46875, 0,6.49351);
\draw[thin,dotted, <->](0, 0,4.46429) -- (0, 0,8.52273);
\draw[thin,dotted, <->](0, 0.681818,6.49351) -- (0, -0.681818,6.49351);

\draw[thick,green!90!black](0.46875,0.46875,5.85938) -- (0.46875,-0.46875,5.85938); 
\draw[thick,green!90!black](0,-0.681818,8.52273) -- (0.46875,-0.46875,5.85938) -- (0,-0.357143,4.46429) --(0,0.357143,4.46429) -- (-0.46875,0.46875,5.85938) -- (0,0.681818,8.52273) -- cycle;
\draw[thick,green!90!black](0,0.357143,4.46429) -- (0.46875,0.46875,5.85938) -- (0,0.681818,8.52273);

\node [circle,fill=black!80,inner sep=0pt,minimum size=3pt,label=south:{}] (pointX) at (0,0,6.49351) {};

\draw[thick,color0](0.46875, -0.681818,8.52273) -- (0.46875, -0.681818,4.46429) -- (0.46875, 0.681818,4.46429) -- (-0.46875, 0.681818,4.46429) -- (-0.46875, 0.681818,8.52273) -- (-0.46875, -0.681818,8.52273) -- (0.46875, -0.681818,8.52273) -- (0.46875, 0.681818,8.52273) -- (-0.46875, 0.681818,8.52273);
\draw[thick,color0](0.46875, 0.681818,8.52273) -- (0.46875, 0.681818,4.46429);

\draw[latex-latex,dashed](-0.56875, 0.781818,8.52273) -- (-0.56875, 0.781818,4.46429); 
\draw[latex-latex,dashed](0.46875, -0.781818,8.62273) -- (-0.46875, -0.781818,8.62273); 
\draw[latex-latex,dashed](0.56875, -0.681818,4.36429) -- (0.56875, 0.681818,4.36429); 

\node [label=center:{$2\varepsilon_{z}$}] at (-0.76875, 0.981818,6.49351) {};
\node [label=center:{$2\varepsilon_{x}$}] at (0, -0.981818,8.82273) {};
\node [label=center:{$2\varepsilon_{y}$}] at (0.76875,0,4.26429) {};

\end{tikzpicture}}  

\caption{Quantization uncertainty bounds for a 3-D point (black). The green shape represents the polyhedral volume and the blue lines represent the cuboid volume.}
\label{fig_bounds}
\end{figure}

In many papers, half the length of a pixel is assumed to be the quantization error for any point measured by the system \cite{schreve_how_2014, chang_quantization_1992, rodriguez_stochastic_1990}, and this was thought to be dependent on depth but independent of other geometric factors such as camera separation, focal length and position in the other two directions. Wu et al. and Fooladgar et al. demonstrated that quantization error is highly dependent on these factors by modelling the uncertainty volumes (see preceding section) rather than just the total magnitude in (\ref{eq:error}). Uncertainty volumes have also been used in other studies as a way to communicate the probabilistic nature of calculated 3-D point positions, for example De Cecco et al. generated uncertainty ellipsoids to take into account the uncertainties from calibration and stereo matching \cite{de_cecco_uncertainty_2009}.

\subsubsection{Occupancy Grid Mapping}\label{sec:occupancy}
In order to deal with sensor uncertainty, occupancy grid mapping represents the environment in a different way to a point cloud. The continuous space of the scene is discretized into a three-dimensional array and one calculates the probability that each gridpoint is occupied. This volumetric reconstruction method came about in the field of autonomous robot navigation with sonar in mind \cite{elfes_using_1989}, and starkly contrasted with surface reconstruction methods in robot perception. Occupancy grids can easily integrate noisy sensor data and different sensor modalities, such as LIDAR and photogrammetry, into a single environment map \cite{pirker_fast_2011}. In addition to modelling objects, the process also models free space and unseen space, which is ideal for applications like path planning and robotic arm manipulation. The accuracy of this technique is dependent on the degree of discretization of the scene space (the resolution of the grid) and the probability model.

Although it has been used with stereo vision and triangulation \cite{moravec_robot_1996}, there has been more work on distance to sensor methods such as disparity maps \cite{lategahn_occupancy_2010, andert_drawing_2009} as these are more computationally efficient for a mobile robot \cite{perrollaz_visibility-based_2012}. 

\subsubsection{Volumetric Scene Representation}
\label{sec_Visual_Hull}
2-D image data has however been used extensively for volumetric scene reconstruction within the field of computer vision.  Early attempts focused on segmenting the scene into a foreground and background and intersecting the foreground silhouettes of multiple views. This represents the maximal space occupied by the object, known as the visual hull, and is represented by voxels \cite{martin_volumetric_1983, laurentini_visual_1994}. One approach to building the visual hull is voxel or space carving, whereby voxels determined to be outside the visual hull are progressively removed from the initial 3-dimensional grid \cite{eisert_automatic_2000, kutulakos_theory_1998, theobalt_combining_2004}. Another approach is to additively build the visual hull using constructive solid geometry intersections \cite{matusik_image-based_2000}.  

Conceptually some of these approaches are similar to the method presented in this paper. Lok describes the visual hull as "the 3-D intersection of all object pixel projection volumes" \cite{lok_online_2001}. The first authors to calculate the pixel viewing a given voxel are Fromhertz et al. in 1994, using the rotation and translation of the object on a rotating stage \cite{fromherz_shape_1994}. Quantization artifacts are a major drawback of volumetric scene reconstruction \cite{seitz_photorealistic_1999, matusik_image-based_2000}, consequently this framework is the ideal way to measure quantization uncertainty. Our method calculates the visual hull of every pixel combination in a system, and from this it would be possible to determine the visual hull of an object spanning multiple pixels in the images. 

\section{Background}\label{sec:background}
\subsection{Pinhole Camera Model}\label{sec:pinhole}
In keeping with earlier studies. we assume the pin-hole camera model for a low distortion system \cite{amat_selection_2002}. Light from a world point at $\vec{X}$, $(X, Y, Z)$ in world coordinates, passes through the aperture or optic center, and is projected to the image at $\vec{w}$, $(u, v)$ in image coordinates. In fig. \ref{fig_perspective}, the optic centers are at $A$ and $B$ and the image planes have been reflected with respect to the $XY$ plane at $z=0$ so that it does not produce a mirror image with negative coordinates, as is more commonly seen in pinhole camera diagrams \cite{sonka_image_2014}. The distance between the cameras or baseline is represented by $b$, with the origin of the world coordinates at the midpoint. For these coplanar cameras, the $x$ and $y$ axes correspond to the $u$ and $v$ directions in image coordinates respectively, and the $z$ axis is parallel to the optical axes, which corresponds to depth in the scene. $f$ is the distance from the optic center to the image plane, commonly referred to as the focal length because this refers to the same quantity in the context of lenses. The image plane is assumed to be the same as the sensor plane, and since the image plane is discretized into pixels, the projected world points will be at within $\pm \frac{1}{2}$ pixel length of the true point \cite{wu_analysis_1998}. 

\subsection{Perspective Projection Model}
The transformation from three-dimensional world coordinates to two-dimensional image coordinates is called perspective projection \cite{cipolla_projection_2019}. Three components are required to model the imaging process from the world point $\vec{X}$ to the pixel $\vec{w}$. These are the pixel to image plane conversion, $\+C$, perspective projection, $\+P$, and a rigid body transformation, $\+R$. These are expressed by the following linear matrix operation in homogeneous coordinates:

\begin{equation}
	\vec{w} = \+C \+P \+R \vec{X}
\end{equation}

\begin{equation}
	\left[\begin{IEEEeqnarraybox*}[][c]{c}
	\IEEEstrut
		su\\sv\\s%
	\IEEEstrut		
	\end{IEEEeqnarraybox*}\right]\!=\!
	\left[\begin{IEEEeqnarraybox*}[][c]{c:c:c}
	\IEEEstrut
		k_u & 0 & u_0 \\
		0 & k_v & v_0 \\
		0 & 0 & 1%
	\IEEEstrut	
	\end{IEEEeqnarraybox*}\right]
	\!
	\left[\begin{IEEEeqnarraybox*}[][c]{c/c/c/c}
	\IEEEstrut	
		f & 0 & 0 & 0 \\
		0 & f & 0 & 0 \\
		0 & 0 & 1 & 0%
		\IEEEstrut
	\end{IEEEeqnarraybox*}\right]\!
	\left[\begin{IEEEeqnarraybox*}[][c]{c/c/c/c}
	\IEEEstrut		
		r_{11} & r_{12} & r_{13} & T_x \\
		r_{21} & r_{22} & r_{23} & T_y \\
		r_{31} & r_{32} & r_{33} & T_z \\
		0 & 0 & 0 & 1%
	\IEEEstrut
	\end{IEEEeqnarraybox*}\right]\!
	\left[\begin{IEEEeqnarraybox*}[][c]{c}
	\IEEEstrut
		\lambda X \\ \lambda Y \\ \lambda Z \\ 1%
	\IEEEstrut
	\end{IEEEeqnarraybox*}\right]\\
\end{equation}	
\begin{equation}
	\left[\begin{IEEEeqnarraybox*}[][c]{c}
	\IEEEstrut
		su \\ sv \\ s%
	\IEEEstrut
	\end{IEEEeqnarraybox*}\right]	
	=
	\left[\begin{IEEEeqnarraybox*}[][c]{c:c:c}
	\IEEEstrut
		a_u & 0 & u_0 \\
		0 & a_v & v_0 \\
		0 & 0 & 1%
	\IEEEstrut
	\end{IEEEeqnarraybox*}\right]
	\left[\begin{IEEEeqnarraybox*}[][c]{c/c/c/c}
	\IEEEstrut		
		r_{11} & r_{12} & r_{13} & T_x \\
		r_{21} & r_{22} & r_{23} & T_y \\
		r_{31} & r_{32} & r_{33} & T_z \\
		0 & 0 & 0 & 1%
	\IEEEstrut
	\end{IEEEeqnarraybox*}\right]
	\left[\begin{IEEEeqnarraybox*}[][c]{c}
	\IEEEstrut	
		\lambda X \\ \lambda Y \\ \lambda Z \\ 1%
	\IEEEstrut	
	\end{IEEEeqnarraybox*}\right]
\end{equation}

where $k$ represents the pixel size in either the $u$ or $v$ direction, $u_{0}$ and $v_{0}$ are the position of the optic axis in the image plane, $f$ is the focal length, $a_{u}=fk_{u}$ and $a_{v}=fk_{v}$ are the image scaling factors, and the ratio $a_{u}/a_{v}$ is the aspect ratio.

In this paper, the optic axis is in the center of the imaging plane, therefore $a_{u}=a_{v}=1$, and we assume square pixels so $k_{u}=k_{v}$. Our pair of cameras are coplanar, meaning their optic axes are parallel so there is no rotational element between them. The world coordinate origin is at the baseline midpoint, therefore the translation $T$ in $x$ will be $T_{x}=-b/2$ for camera A and $T_{x}=b/2$ for camera B, with $T_{y}=T_{z}=0$.

\section{Methodology}

\begin{figure}[!t]
\centering
     \scalebox{1}{\begin{tikzpicture}
\node [circle, fill=black,inner sep=0pt,minimum size=4pt,label=south:{A}] (v2) at (-1.5,-1) {};

\node (v4) at (-2.2,1) {};
\draw (-2.2,1.1) -- (-2.2,0.9);
\node (v5) at (-1.5,) {};
\draw (-1.5,1.1) -- (-1.5,0.9);
\node (v3) at (-2.9,1) {};
\draw (-2.9,1.1) -- (-2.9,0.9);
\node (v1) at (-3.6,1) {};
\draw (-3.6,1.1) -- (-3.6,0.9); 

\draw  [thick] (v1.center) edge (v5.center);
\draw  (v1.center) edge (v2.center);
\draw  (v3.center) edge (v2.center);
\draw  (v4.center) edge (v2.center);
\draw  [dotted](v5.center) edge (v2.center);


\draw  (v1.center) edge (-4.6,2);
\draw  (v3.center) edge (-3.95,2.5);
\draw  (v4.center) edge (-2.9,3);
\draw [dotted] (v5.center) edge (-1.5,3.5);

\draw [very thin] (-1.5,3) arc (90:109.4:4); 
\draw [very thin] (-1.5,2.6) arc (90:125.1:3.6);
\draw [very thin] (-1.5,2.2) arc (90:136.4:3.2);
%

%
\node [label=center:{$\alpha_{1}$}] (v10) at (-2.2,3.1) {};
\node [label=center:{$\alpha_{2}$}] (v11) at (-3.3,2.35) {};
\node [label=center:{$\alpha_{3}$}] (v12) at (-3.75,1.55) {};
\node [label=center:{k}] (v13) at (-3.1,0.8) {};
\node [label=center:{k}] (v14) at (-2.5,0.8) {};
\node [label=center:{k}] (v15) at (-1.8,0.8) {};
\draw [latex-latex, dashed] (-1.3,-1) to (-1.3,1){};
\node [label=center:{$f$}](v16) at (-1.1,0) {};

\end{tikzpicture}}  

\caption{The angles subtended by each pixel, $k$, vary as a function of distance from the optic center (dotted line).}
\label{fig_pixel_angles}
\end{figure}
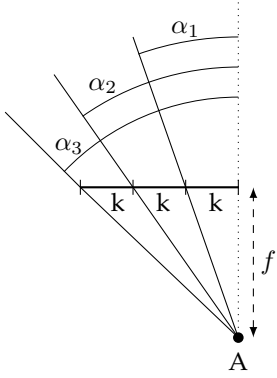

In this section, we summarize how we calculate all the polyhedron uncertainty volumes within a region of space represented by an occupancy grid, which involves simple vector geometry. We then detail how the results can be visualized and queried, such as determining the volume for a specific pair of pixels obtained during correspondence matching. We also present the error metrics used to measure the accuracy of our results.

\subsection{Computation of the Intersection Region}
We start by defining our region of interest, this can be the entire volume where the two camera views intersect, or a smaller subset of that space. We generate regularly spaced gridpoints filling this volume with cartesian coordinates $(X, Y, Z)$. The origin of the coordinate system is at the baseline midpoint. The x-axis is parallel to the baseline, the z-axis is depth and the y-axis is perpendicular to these directions. These grid points can be thought of as the centroids or vertices of cube shaped voxels. The accuracy of this method is highly dependent on the spacing between grid points, so this should be chosen carefully. In section \ref{sec:gridpoints} we investigate how the grid point spacings affect the calculated volumes for the system we model and a suitable spacing is chosen for the following experiments. 

The following steps, summarized in algorithm 1, determine which gridpoints are in the view of each pixel in one camera. This process is repeated for every other camera or view in the system. We start by specifying the following parameters for the camera:
\begin{itemize}
\item Perspective center in world coordinates, $A$ and $B$
\item Angular field of view in horizontal and vertical directions, $FOV$
\item Focal length, $f$
\item Number of pixels, $n$
\item Pixel size, $k$
\end{itemize}

In this study we assume square pixels and equal numbers of detectors in the rows and columns of the sensor. We assume each pixel subtends an angle bounded by planes passing through the camera perspective center, $A$, and each pixel border, as shown in fig. \ref{fig_pixel_angles} in 2-D. The pixel border angles, $\alpha$, are calculated using perspective projection, consequently the angle viewed by each pixel decreases with increasing distance from the optic axis. 

To calculate the $\alpha$ angles, we generate a vector containing the x coordinates of the edges of each pixel $\vec{x_{k}}$, defined by the following sequence $\langle x \rangle$: 

\begin{equation}
\langle x \rangle = ki, \quad i \in \mathbb{Z}, \quad \frac{-n}{2} \leq i \leq \frac{n}{2}
\end{equation}

CCDs always have dimensions of powers of 2, so $n$ will always be even, $n=2x,x \in \mathbb{N}$. $\alpha$ is then the inverse tangent of $\langle x \rangle$ divided by the focal length. 

\begin{equation}
\vec{\alpha}=\arctan\left(\frac{1}{f}.\vec{x_{k}}\right)
\label{eq:alpha}
\end{equation}

Now that we have the horizontal angles for the pixel edges, we perform the same procedure to determine the vertical angles for the pixel edges, $\gamma$. In this work we simulate square pixels therefore $\gamma=\alpha$. Next we convert the Cartesian coordinates of each scene grid point, $P$, to spherical coordinates with an origin also at the camera center, using the following:

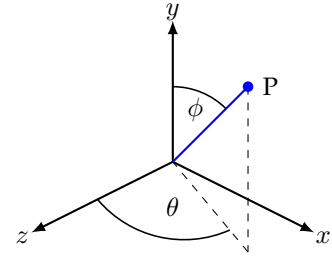
\begin{figure}[!t]
\centering
     \scalebox{1}{\begin{tikzpicture}
\node (v1) at (0,0) {};
\node [label=center:{$y$}](v4) at (0,2) {};
\node [label=center:{$z$}](v2) at (-2,-1) {};
\node [label=center:{$x$}](v3) at (2,-1) {};

\draw [-latex,thick](v1.center) -- +(v2);
\draw [-latex,thick](v1.center) -- +(v3);
\draw [-latex,thick] (v1.center) -- +(v4);

\node (v5) at (1,1) {};
\node (v6) at (1,-1.2) {};
\draw [blue,thick] (v1.center) edge (v5.center);
\draw [dashed](v5.center) edge (v6.center);
\draw [dashed](v6.center) edge (v1.center);
\node [circle, fill=blue,inner sep=0pt,minimum size=4pt,label=east:{P}] (75) at (1,1) {};

\draw [semithick] (0,1) arc (90:45:1);
\node at (0.3,0.7) {$\phi$};


\node at (0,-0.6) {$\theta$};
\draw [semithick](-1,-0.5) arc (-140.:-66:1.5);

\end{tikzpicture}}  

\caption{Spherical coordinates used to define the position of point P.}
\label{fig_spherical}
\end{figure}

\begin{IEEEeqnarray}{rCl}
	\rho &=& \sqrt{x^{2}+y^{2}+z^{2}}\\
	\phi &=& \arctan \left(y/z\right)\\
	\theta &=& \arctan \left(x/z\right)%
\end{IEEEeqnarray}

whereby $x$, $y$ and $z$ are the cartesian coordinates, $\rho$ is the distance from the optic center to $P$, $\phi$ is the elevation, and $\theta$ is the azimuth relative to the optic axis, shown in fig. \ref{fig_spherical}.

We then determine which pixel view each grid point appears in. The angles in $\alpha$ are ordered and increasing, therefore, we can determine where $\theta$ for each grid point fits in to $\alpha$ by performing a binary search. This returns the index of the element in $\alpha$ such that if $\theta$ were inserted into $\alpha$, the order of $\alpha$ would be preserved \cite{van_der_walt_numpy_2011}. This index is $u$ in the case of $\theta$ and $\alpha$, and $v$ in the case of $\phi$ and $\gamma$. These are the pixel ID, $(u,v)$, for the simulated camera, shown in fig. \ref{fig_perspective}.

We now know for every grid point the cartesian coordinates, spherical coordinates, and pixel ID. These values are stored in a lookup table, $LUT$, with the pixel ID as the keys and coordinates as the values. This allows us to look up all the gridpoints seen by a particular pixel, but the keys and values can be readily inverted so that we can look up the pixel viewing each gridpoint. The algorithm is summarized in algorithm 1.

We repeat this process for every other camera in the system, giving us a lookup table for each one, in our case $LUT_{A}$ and $LUT_{B}$. By performing an intersection of the gridpoint coordinates in these lookup tables (intersection used in terms of set theory rather than geometry) we obtain both pixel IDs for each gridpoint. This new lookup table, $LUT_{AB}$, has the pixel IDs of both cameras as the key, $[(u_A,v_A ),(u_B,v_B )]$, and all the gridpoints within that intersection as the values. The algorithm steps are described in algorithm 2. If there are three or more views in the system you repeat algorithm 2 with $LUT_{AB}$ and $LUT_{C}$ and so on. 

$LUT_{AB}$ contains all the information needed to determine which pixels in cameras $A$ and $B$ see any scene point, and conversely which scene points are in the uncertainty volume of any pixel combination.

\algsetup{indent=2em}
\noindent\rule{\linewidth}{1pt}
\textbf{Algorithm 1:} Gridpoints in view of one camera
\vspace{1mm}\hrule\vspace{1mm}
\noindent\textbf{Input:} $R$ = region of interest, $s$ = gridpoint spacing, camera parameters$[A, f, n, k]$ 
\begin{algorithmic}
\STATE $G$ = GenerateOccupancyGrid$(R, s)$
\STATE $A$ = new array of length $i$
	\FOR{$x$ in $\langle x \rangle$, where $\langle x_{i} \rangle$ is defined in equation 5}
		\STATE compute $\alpha$ using equation \ref{eq:alpha}
		\STATE $A[i] = \alpha$
	\ENDFOR
\STATE $S$ = new 3-D array with dimensions $G$
	\FOR{$x,y,z$ point in $G$}
		\STATE compute $\rho,\theta,\phi$ using equations 7, 8, and 9
		\STATE $S[x, y, z] = \rho,\theta,\phi$
	\ENDFOR
\STATE $U$ = new 3-D array with dimensions $G$
\STATE $V$ = new 3-D array with dimensions $G$
	\FOR{$\rho,\theta,\phi$ in $S$ with index $x, y, z$:}
		\STATE $U[x, y, z]$ = FindBinarySearch$(A,\theta)$
		\STATE $V[x, y, z]$ = FindBinarySearch$(A,\phi)$
	\ENDFOR
\STATE $LUT$ = new lookup table
	\FOR{$x, y, z$ in $G$}
		\STATE $u = U[x, y, z]$
		\STATE $v = V[x, y, z]$
		\STATE add point $P(x, y, z)$ to current list in $R[u, v]$ 
	\ENDFOR
\end{algorithmic}
\noindent\textbf{Output:} $LUT$

\noindent\rule{\linewidth}{1pt}
\textbf{Algorithm 2:} Gridpoints in view of 2 or more cameras
\vspace{1mm}\hrule\vspace{1mm}
\noindent\textbf{Input:} $LUT_{A}$, $LUT_{B}$ 
\begin{algorithmic}
\STATE $LUT_{AB}$ = new lookup table
	\FOR{$u_{A}$, $v_{A}$ in $LUT_{A}$:}
		\FOR{$u_{B}$, $v_{B}$ in $LUT_{B}$:}
			\STATE $points_{A} = LUT_{A}[u_{A}, v_{A}]$
			\STATE $points_{B} = LUT_{B}[u_{B}, v_{B}]$
			\STATE intersection = $points_{A} \cap points_{B}$
				\IF {intersection:}
					\STATE intersections$[u_{A},v_{A},u_{B},v_{B}]$ = intersection
				\ENDIF
				\ENDFOR
				\ENDFOR
\end{algorithmic}
\noindent\textbf{Output:} $LUT_{AB}$\\
\rule{\linewidth}{0.4pt}

\subsection{Uncertainty Region Representation}
The lookup table made in algorithm 2 can be queried to characterize and plot the uncertainty volumes for cameras $A$ and $B$. The polyhedron volume is obtained by counting the number of points, $p$, in each intersection and converting this to a volume in meters cubed using the spacing between grid points, $g$:

\begin{equation}
V = g^{3}p
\label{eq:volume}
\end{equation}

The cuboid uncertainty volume and its dimensions offer an approximate shape of this region as shown in fig. \ref{fig_bounds}. The cuboid volume is calculated by querying the minimum and maximum coordinate values in an intersection, calculating the $x$, $y$, and $z$ dimensions and multiplying these together. It is possible to examine how the three components contribute to the volume as we vary parameters such as distance from the camera, however the usefulness of this is limited by the degree to which the major and minor axes of the polyhedron align with the cartesian axes. 

\subsection{Error Metrics}
To test the validity of our approach we compare our results with those of previous authors in terms of Root Mean Squared (RMS) error:

\begin{equation}
RMS = \sqrt{\frac{\sum_{n=1}^{N} (p-o)^{2}}{N}}
\end{equation} 

where $p$ is the predicted value, $o$ is the observed value and $N$ is the number of measurements. For our calculations the predicted values are the cuboid volumes obtained by \cite{fooladgar_geometrical_2013} and the observed values are the cuboid volumes obtained using our method. We want to express the error as a percentage, so rather than use the Mean Absolute Percentage Error (MAPE) \cite{de_myttenaere_mean_2016}, which imposes a heavier penalty when $o<p$, we use a logarithm of the accuracy ratio, $p/o$. This outperforms MAPE when the error grows as a function of a variable \cite{tofallis_better_2014}, which is the case for all our observations. We calculate the Median Symmetric Accuracy (MSA) \cite{morley_measures_2018}, given by the following:

\begin{equation}
MSA = 100 ((e^{M|log_{e}(p/o)|})-1)
\end{equation}

where $M$ represents the median function of all values of $log_{e}(p/o)$.

\section{Results}
\begin{figure*}[!t]
\centering
\subfloat[]{
\begin{tikzpicture}
\tikzstyle{every node}=[font=\scriptsize]
\definecolor{color0}{rgb}{0,0.469313725490196,0.8667}
\definecolor{color1}{rgb}{1,0.788235294117647,0}

\begin{axis}[
ticklabel style={
/pgf/number format/fixed,
/pgf/number format/precision=5
}, scaled ticks=false, 
legend cell align={left},
legend style={fill opacity=0.8, draw opacity=1, text opacity=1, draw=white!80!black},
tick align=outside,
tick pos=both,
x grid style={white!95!black},
xlabel={Baseline Distance (m)},
xmajorgrids,
xmin=0.0615680216499994, xmax=104.75897295135,
xtick style={color=black},
y grid style={white!95!black},
ylabel={Volume of Uncertainty Region (m\textsuperscript{3})},
ymajorgrids,
ymin=-0.00270337340764785, ymax=0.108233528447983,
ytick style={color=black}
]
\addplot [line width=0.56pt, black, mark=*, mark size=2, mark options={solid}, only marks]
table {%
4.820540973 0.103190942
9.945453759 0.048036577
14.96357815 0.031594342
20.03675927 0.023458749
24.88853853 0.019161242
29.89834518 0.016063807
34.95543272 0.013617944
40.00209044 0.011923644
44.86185225 0.010874463
49.90457518 0.009809579
54.9275349 0.008740918
60.0166316 0.007982629
64.81548543 0.007325199
69.94159198 0.006963017
74.93712985 0.006329506
79.98509917 0.006029619
84.8531308 0.005578901
89.88384923 0.005624361
94.93181855 0.005332888
99.95256613 0.005014515
};
\addlegendentry{Fooladgar et al.}
\addplot [line width=0.56pt, color0, mark=*, mark size=2, mark options={solid}, only marks]
table {%
5 0.096174
10 0.04677075
15 0.029744
20 0.02197
25 0.018083
30 0.0149565
35 0.0127595
40 0.010985
45 0.009971
50 0.008957
55 0.008112
60 0.007267
65 0.0068445
70 0.006422
75 0.005915
80 0.005408
85 0.005239
90 0.004901
95 0.004732
100 0.004394
};
\addlegendentry{Cuboid Volume}
\addplot [line width=0.56pt, green!73.853137254902!black, mark=*, mark size=2, mark options={solid}, only marks]
table {%
5 0.0498343551252257
10 0.0234481477070222
15 0.0154197840209849
20 0.0117458986735427
25 0.00946453438521168
30 0.00781523639321479
35 0.00664986415934873
40 0.00589928543245194
45 0.005231007004557
50 0.00471745313878551
55 0.00423681939261476
60 0.00392078624444769
65 0.00361462913216084
70 0.00334139213947473
75 0.00311753532618973
80 0.00295293472818604
85 0.00276199803450177
90 0.00259739743649809
95 0.00245254891025485
100 0.00239987671889368
};
\addlegendentry{Polyhedron Volume}
\addplot [line width=1.5pt, color1, dashed]
table {%
5 0.0479190293035821
6 0.0398743223957534
7 0.0341358681192017
8 0.0298369941707713
9 0.0264967938472818
10 0.023827022577095
11 0.0216444198537632
12 0.0198269120593175
13 0.018290049346643
14 0.0169735512631545
15 0.0158332397213866
16 0.0148360003128543
17 0.0139565238985258
18 0.0131751355233029
19 0.0124763096954324
20 0.0118476315806121
21 0.011279055333213
22 0.0107623649314029
23 0.0102907758255825
24 0.00985863628994923
25 0.00946120053315849
26 0.00909445422939873
27 0.00875497886600162
28 0.00843984519382637
29 0.00814652874860375
30 0.00787284228818537
31 0.00761688132212997
32 0.00737697986678005
33 0.00715167425458163
34 0.00693967333772937
35 0.00673983380596308
36 0.00655113962307573
37 0.00637268480211468
38 0.00620365890361796
39 0.00604333476762936
40 0.0058910580881747
41 0.00574623851530519
42 0.00560834202984741
43 0.00547688438345838
44 0.00535142543432744
45 0.00523156423905593
46 0.00511693478552262
47 0.0050072022711686
48 0.00490205984707684
49 0.00480122576123415
50 0.00470444084503061
51 0.00461146629583545
52 0.00452208171575049
53 0.00443608337267043
54 0.00435328265480153
55 0.00427350469399057
56 0.00419658713674017
57 0.00412237904475402
58 0.00405073990936258
59 0.00398153876630327
60 0.00391465339913466
61 0.00384996962110259
62 0.00378738062659122
63 0.00372678640441898
64 0.00366809320620783
65 0.00361121306388854
66 0.00355606335112484
67 0.00350256638406301
68 0.00345064905735437
69 0.00340024251186818
70 0.00335128183092277
71 0.00330370576221991
72 0.00325745646298095
73 0.00321247926605748
74 0.00316872246503011
75 0.00312613711652148
76 0.00308467685813602
77 0.00304429774060441
78 0.00300495807285654
79 0.0029666182788762
80 0.0029292407653054
81 0.00289278979886837
82 0.00285723139277604
83 0.00282253320135283
84 0.00278866442219972
85 0.00275559570527237
86 0.00272329906831077
87 0.00269174781810883
88 0.00266091647715887
89 0.00263078071524779
90 0.00260131728561935
91 0.00257250396535103
92 0.00254431949962426
93 0.00251674354959482
94 0.00248975664359497
95 0.00246334013142147
96 0.00243747614148435
97 0.00241214754060963
98 0.00238733789630635
99 0.00236303144132335
100 0.00233921304033538
};
\addlegendentry{Power law fit}
\end{axis}

\end{tikzpicture}
\label{fig_baseline}}
\hfil
\subfloat[]{\input{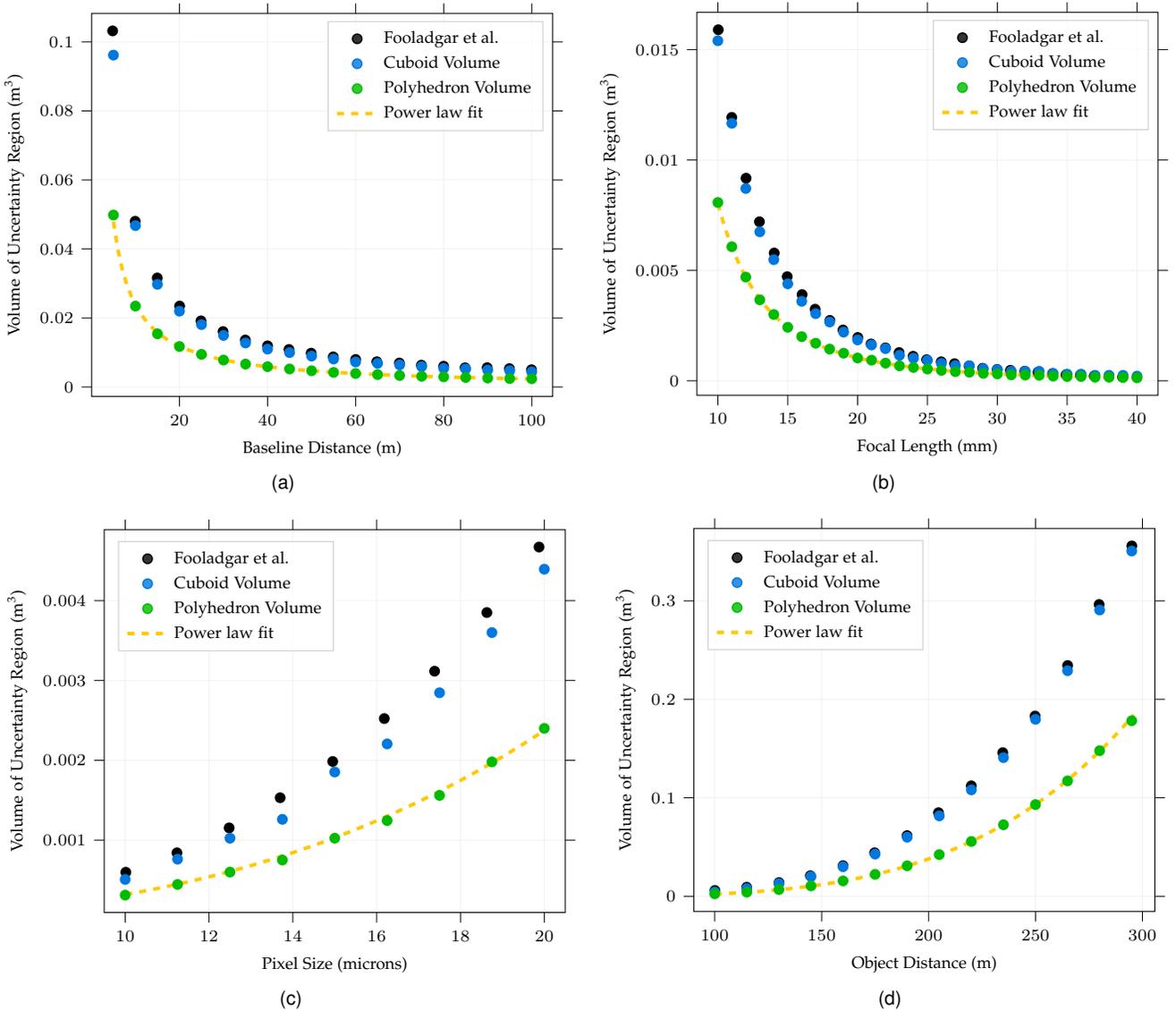}
\label{fig_focal}}
\vfil
\subfloat[]{
\begin{tikzpicture}
\tikzstyle{every node}=[font=\scriptsize]
\definecolor{color0}{rgb}{0,0.469313725490196,0.8667}
\definecolor{color1}{rgb}{1,0.788235294117647,0}

\begin{axis}[
ticklabel style={
/pgf/number format/fixed,
/pgf/number format/precision=5
}, scaled ticks=false, 
legend cell align={left},
legend style={fill opacity=0.8, draw opacity=1, text opacity=1, at={(0.03,0.97)}, anchor=north west, draw=white!80!black},
tick align=outside,
tick pos=both,
x grid style={white!95!black},
xlabel={Pixel Size (microns)},
xmajorgrids,
xmin=9.5, xmax=20.5,
xtick style={color=black},
y grid style={white!95!black},
ylabel={Volume of Uncertainty Region (m\textsuperscript{3})},
ymajorgrids,
ymin=9.26114073326291e-05, ymax=0.00488986498060321,
ytick style={color=black}
]
\addplot [line width=0.56pt, black, mark=*, mark size=2, mark options={solid}, only marks]
table {%
10.01488334 0.000597171
11.23552878 0.00084099
12.48006447 0.001152867
13.69695023 0.001530406
14.95149855 0.001984676
16.18252588 0.002522795
17.38160149 0.003116243
18.62923202 0.003850716
19.8746054 0.004671808
};
\addlegendentry{Fooladgar et al.}
\addplot [line width=0.56pt, color0, mark=*, mark size=2, mark options={solid}, only marks]
table {%
10 0.000507
11.25 0.00076125
12.5 0.001024
13.75 0.00126
15 0.0018525
16.25 0.002205
17.5 0.00284625
18.75 0.0036
20 0.004394
};
\addlegendentry{Cuboid Volume}
\addplot [line width=0.56pt, green!73.853137254902!black, mark=*, mark size=2, mark options={solid}, only marks]
table {%
10 0.000310668387935837
11.25 0.00044442161460994
12.5 0.000599024250364509
13.75 0.000750578726896787
15 0.00102418149868957
16.25 0.00124462437364561
17.5 0.00155968211086155
18.75 0.00197971845169317
20 0.00239987671889368
};
\addlegendentry{Polyhedron Volume}
\addplot [line width=1.5pt, color1, dashed]
table {%
10 0.000317731293889713
10.5 0.000366023490816292
11 0.000418888828625561
11.5 0.000476522847580226
12 0.000539120200484254
12.5 0.000606874695090417
13 0.000679979332798096
13.5 0.000758626344101512
14 0.000843007221175874
14.5 0.000933312747930103
15 0.00102973302780661
15.5 0.00113245750956897
16 0.00124167501128532
16.5 0.00135757374268802
17 0.00148034132606669
17.5 0.00161016481583243
18 0.00174723071687442
18.5 0.00189172500181569
19 0.00204383312726293
19.5 0.00220374004913456
20 0.00237163023714223
};
\addlegendentry{Power law fit}
\end{axis}

\end{tikzpicture}
\label{fig_pixel}}
\hfil
\subfloat[]{
\begin{tikzpicture}
\tikzstyle{every node}=[font=\scriptsize]

\definecolor{color0}{rgb}{0,0.469313725490196,0.8667}
\definecolor{color1}{rgb}{1,0.788235294117647,0}

\begin{axis}[
legend cell align={left},
legend style={fill opacity=0.8, draw opacity=1, text opacity=1, at={(0.03,0.97)}, anchor=north west, draw=white!80!black},
tick align=outside,
tick pos=both,
x grid style={white!95!black},
xlabel={Object Distance (m)},
xmajorgrids,
xmin=90.2, xmax=305.8,
xtick style={color=black},
y grid style={white!95!black},
ylabel={Volume of Uncertainty Region (m\textsuperscript{3})},
ymajorgrids,
ymin=-0.0153535949569425, ymax=0.373435183188426,
ytick style={color=black}
]
\addplot [line width=0.56pt, black, mark=*, mark size=2, mark options={solid}, only marks]
table {%
100.0881184 0.00589539
114.9451109 0.009156124
130.0553725 0.013880738
144.7082345 0.020853153
160.0072916 0.031093227
174.7685114 0.044196775
189.9476165 0.06159323
204.6367144 0.084888441
219.9685431 0.112106913
234.6814559 0.145837341
249.8026288 0.183041314
265.0264374 0.234338395
279.7701918 0.29625298
295.0479144 0.355762966
};
\addlegendentry{Fooladgar et al.}
\addplot [line width=0.56pt, color0, mark=*, mark size=2, mark options={solid}, only marks]
table {%
100 0.004394
115 0.007875
130 0.0128605
145 0.0200355
160 0.029988
175 0.042849
190 0.06
205 0.081648
220 0.1080685
235 0.1407865
250 0.179685
265 0.229075
280 0.29068125
295 0.3505905
};
\addlegendentry{Cuboid Volume}
\addplot [line width=0.56pt, green!73.853137254902!black, mark=*, mark size=2, mark options={solid}, only marks]
table {%
100 0.00245408784376047
115 0.0041627105034614
130 0.00682937873504951
145 0.0105614425173633
160 0.015606642161613
175 0.0222796714501312
190 0.0309605567888409
205 0.0423240694732982
220 0.0556713748561283
235 0.0726183332459801
250 0.093100858825057
265 0.117187027189695
280 0.147836879974744
295 0.17824136137959
};
\addlegendentry{Polyhedron Volume}
\addplot [line width=1.5pt, color1, dashed]
table {%
100 0.00231862223148331
102 0.00251134713678773
104 0.00271587716173481
106 0.00293268741159828
108 0.00316226251605099
110 0.00340509663499102
112 0.00366169346426236
114 0.00393256624127387
116 0.0042182377505203
118 0.00451924032900859
120 0.00483611587159278
122 0.00516941583622071
124 0.00551970124909516
126 0.00588754270975254
128 0.00627352039606151
130 0.00667822406914422
132 0.00710225307822242
134 0.00754621636539091
136 0.00801073247032035
138 0.00849642953489164
140 0.00900394530776379
142 0.00953392714887721
144 0.0100870320338943
146 0.010663926558579
148 0.0112652869431171
150 0.0118917990363788
152 0.0125441583201251
154 0.0132230699131595
156 0.0139292485754266
158 0.0146634187120586
160 0.0154263143773717
162 0.0162186792788124
164 0.0170412667808565
166 0.0178948399088609
168 0.0187801713528698
170 0.0196980434713756
172 0.0206492482950374
174 0.0216345875303557
176 0.0226548725633058
178 0.0237109244629314
180 0.0248035739848971
182 0.025933661575003
184 0.0271020373726606
186 0.0283095612143325
188 0.0295571026369344
190 0.0308455408812025
192 0.0321757648950259
194 0.0335486733367441
196 0.0349651745784122
198 0.0364261867090323
200 0.0379326375377536
202 0.0394854645970406
204 0.0410856151458108
206 0.0427340461725416
208 0.0444317243983481
210 0.0461796262800321
212 0.0479787380131014
214 0.0498300555347623
216 0.0517345845268845
218 0.0536933404189381
220 0.0557073483909053
222 0.0577776433761655
224 0.0599052700643551
226 0.062091282904202
228 0.0643367461063359
230 0.066642733646074
232 0.0690103292661832
234 0.0714406264796188
236 0.0739347285722404
238 0.0764937486055052
240 0.0791188094191388
242 0.0818110436337846
244 0.0845715936536318
246 0.0874016116690214
248 0.0903022596590329
250 0.0932747093940493
252 0.0963201424383022
254 0.099439750152398
256 0.102634733695824
258 0.105906304029433
260 0.109255681917916
262 0.112684097932247
264 0.116192792452117
266 0.119783015668346
268 0.123456027585276
270 0.127213098023154
272 0.13105550662049
274 0.134984542836397
276 0.139001505952924
278 0.143107705077363
280 0.147304459144542
282 0.151593096919108
284 0.155974956997781
286 0.160451387811609
288 0.165023747628194
290 0.169693404553911
292 0.174461736536107
294 0.179330131365289
296 0.184299986677293
};
\addlegendentry{Power law fit}
\end{axis}

\end{tikzpicture}
\label{fig_object}}
\caption{Volume of uncertainty region versus (a) baseline distance, (b) focal length, (c) pixel size, (d) object distance in the $z$ direction.}
\label{fig_experiments}
\end{figure*}
We present a series of experiments to compare the uncertainty regions calculated using this method with those obtained by earlier studies \cite{fooladgar_geometrical_2013,wu_analysis_1998}. Fooladgar et al. showed that the pyramids intersection method used by Wu et al., produced error measures “almost equal” to their cone intersection method. We ran our simulation at “a sufficiently high resolution”, discussed in section \ref{sec:gridpoints}, and calculated the polyhedron error volumes and the cuboid error volumes. We present these two measures alongside those obtained by \cite{fooladgar_geometrical_2013}, and we compare our cuboid volumes with their cuboid volumes in terms of Root Mean Square Error (RMSE) and Median Symmetric Accuracy (MSA), shown in table \ref{table}. Point for point comparison was not possible as their original data was unavailable so the process of extracting it from the figures introduced a precision error in their results. Our cuboid volumes are consistently lower than theirs because we quantize our scene.

For each experiment, unless the parameter is being varied, the baseline is 100 m, the focal length 15 mm, the pixel size 20 $\mu$m, as these were used in the earlier study \cite{fooladgar_geometrical_2013}. The uncertainty volume for an object 100 m away from the cameras was measured each time, except for the object distance experiment where this was increased up to 300 m. The vertical axes for fig \ref{fig_experiments} show the volume of the uncertainty region in meters cubed, and the horizontal axis the variable of each experiment: baseline, focal length, pixel size, object distance, $z$. The gridpoint spacing used to find the cuboid volume and polyhedron volume is $3$ per cm, this is further discussed in section \ref{sec:gridpoints}.

\subsection{Error due to system parameters}

Figure \ref{fig_baseline} shows the effect of baseline length, or camera separation, on uncertainty volume. The decrease in polyhedron volume with increasing baseline distance follows a power law of $V=0.2427b^{-1.008}$, where $V$ is the volume of the uncertainty region and $b$ is the baseline distance. Earlier authors reported a similar trend and the results of \cite{fooladgar_geometrical_2013} are plotted alongside ours for comparison. They attributed the decrease in uncertainty volume with increasing baseline distance to the viewing pixel being further from the sensor center. As the pixel viewing angles decrease the pyramid projections narrow, resulting in smaller intersection volumes, and our findings confirm this. For smaller baselines, the difference between polyhedron volume and cuboid volume is at its largest, and for larger baselines the difference is less marked. The cuboid volume calculation overestimates the localization error by almost two times at 5 m baseline, therefore for smaller baselines the polyhedron volume model would be more suitable. 

Figure \ref{fig_focal} shows the effect of focal length on error volume. The longer the focal length, the narrower the pyramid so the pyramid intersections are smaller in these cases. The decrease in polyhedron volume also follows a power law of $V=7.1298f^{-2.948}$, and the difference between the two models of error volume is greatest for smaller focal lengths. Figure \ref{fig_pixel} shows the effect of pixel size on error volume. The increase in polyhedron volume as a function of pixel size can be described with the following power law, $V=4\times10^{-7}k^{2.9253}$. The larger the pixel, the wider the pyramid cross section, so the uncertainty volume increases with increasing pixel size. 

\begin{table}[!t]
\renewcommand{\arraystretch}{1.3}
\caption{Root mean squared Error (RMSE) and median symmetric accuracy (MSA) of cuboid volumes obtained using our method compared with cuboid volumes obtained by Fooladgar et al.}
\label{table}
\centering
\begin{IEEEeqnarraybox}[
\IEEEeqnarraystrutmode
\IEEEeqnarraystrutsizeadd{3pt}
{1pt}
]{.u/t/t/t/t.}
\hline
& Baseline & Focal Length & Pixel Size & Object Distance\\ \hline
RMSE (m\textsuperscript{3}) & 0.001808 & 0.000196 & 0.00022 & 0.00343 \\ 
MSA (\%) & 7.58 & 3.07 & 10.47 & 2.30 \\ \hline
\end{IEEEeqnarraybox}
\end{table}

\subsection{Error due to scene point position}

Figure \ref{fig_object} shows the effect of object distance or increasing depth, $z$, on the error volume. The point to be imaged is 100 m away from the baseline midpoint (0, 0, 100), and the $z$ value is increased up to 300 m. The uncertainty volume increase with object distance follows a power law of $V=2 \times 10^{-11} z^{4.0321}$. The further away the object, the closer to the center of the sensor it is detected, and the wider the viewing pyramids, hence the larger the intersection volume.

\begin{figure}[!t]
\centering
\includegraphics[width=3.5in, trim=120 25 70 60, clip]{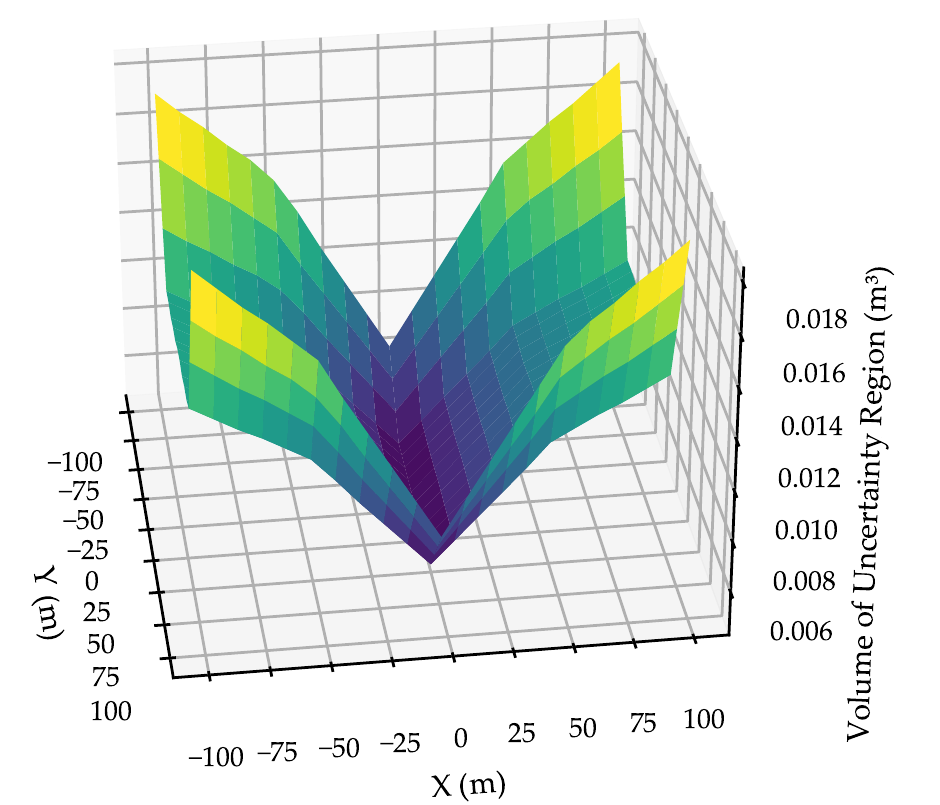}
\caption{Volume of cuboid uncertainty region versus object positions within the $XY$ plane at $z = 100\,m$.}
\label{fig_plane}
\end{figure}

We now look beyond the $z$ axis to measure uncertainty volumes at different $x$ and $y$ positions within a plane perpendicular to this axis. Fig. \ref{fig_plane} shows the effect of the $x$ and $y$ coordinate on the cuboid error volume. At $z = 100$ m, a plane of scene points from -100 m to +100 m in $x$ and $y$ was generated, and the error volume for each point in this plane was calculated. In the $x$ direction the uncertainty volumes are lowest at the CCD center and increase as we move left or right in the plane. The cameras are situated at (50, 0, 0) and (-50, 0, 0) in m, and we observe a smaller increase in uncertainty volume for scene points beyond theses $x$ values. The error volume does not change with variation in the $y$ direction for points within $\pm 50$ m in $y$ position but it increases beyond these values. The extent of the same simulations in \cite{fooladgar_geometrical_2013} and \cite{wu_analysis_1998} was $\pm 50$ m in $x$ and $y$ rather than $\pm 100$ m. Our results closely match theirs for the same region and we show the trends in uncertainty volumes change beyond this region. We build on earlier studies by also examining how the uncertainty volumes change as a function of position in the other two perpendicular planes, $XZ$ and $YZ$.

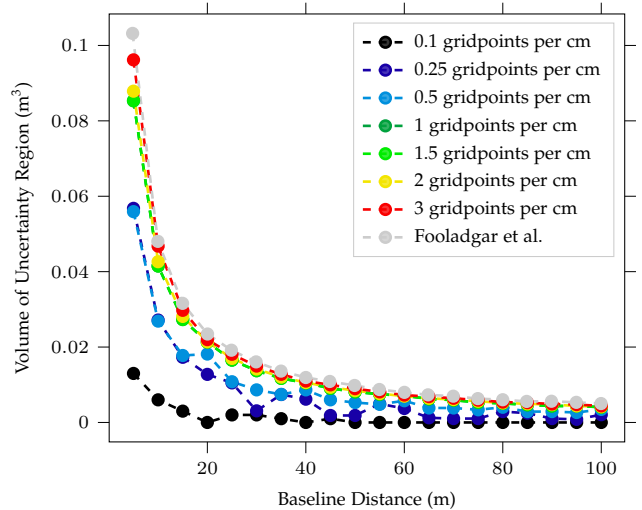
\begin{figure}[!t]
\centering
\begin{tikzpicture}
\tikzstyle{every node}=[font=\scriptsize]
\definecolor{color0}{rgb}{0.0941117647058824,0,0.654929411764706}
\definecolor{color1}{rgb}{0,0.563407843137255,0.8667}
\definecolor{color2}{rgb}{0,0.630080392156863,0.240507843137255}
\definecolor{color3}{rgb}{0.951609803921569,0.896707843137255,0}

\begin{axis}[
ticklabel style={
/pgf/number format/fixed,
/pgf/number format/precision=5
}, scaled ticks=false, 
legend cell align={left},
legend style={fill opacity=0.8, draw opacity=1, text opacity=1, draw=white!80!black},
tick align=outside,
tick pos=both,
x grid style={white!69.0196078431373!black},
xlabel={Baseline Distance (m)},
xmin=0.0615680216499994, xmax=104.75897295135,
xtick style={color=black},
y grid style={white!69.0196078431373!black},
ylabel={Volume of Uncertainty Region (m\textsuperscript{3})},
ymin=-0.0051595471, ymax=0.1083504891,
ytick style={color=black}
]
\addplot [line width=1pt, black, dashed, mark=*, mark size=2, mark options={solid}]
table {%
5 0.013
10 0.006
15 0.003
20 0
25 0.002
30 0.002
35 0.001
40 0
45 0.001
50 0
55 0
60 0
65 0
70 0
75 0
80 0
85 0
90 0
95 0
100 0
};
\addlegendentry{0.1 gridpoints per cm}
\addplot [line width=1pt, color0, dashed, mark=*, mark size=2, mark options={solid}]
table {%
5 0.05678976
10 0.02716032
15 0.01728384
20 0.01275216
25 0.01049376
30 0.0030816
35 0.00740736
40 0.0061704
45 0.00184896
50 0.00184896
55 0.00493824
60 0.00370224
65 0.00123264
70 0.0010272
75 0.0010272
80 0.00287952
85 0.00246912
90 0.0010272
95 0.00082176
100 0.0020568
};
\addlegendentry{0.25 gridpoints per cm}
\addplot [line width=1pt, color1, dashed, mark=*, mark size=2, mark options={solid}]
table {%
5 0.05592
10 0.02688
15 0.01776
20 0.018144
25 0.0108
30 0.00864
35 0.00744
40 0.00864
45 0.006
50 0.00528
55 0.0048
60 0.00576
65 0.00384
70 0.00384
75 0.00336
80 0.004032
85 0.00288
90 0.00288
95 0.00264
100 0.003456
};
\addlegendentry{0.5 gridpoints per cm}
\addplot [line width=1pt, color2, dashed, mark=*, mark size=2, mark options={solid}]
table {%
5 0.085332
10 0.041496
15 0.0273
20 0.021294
25 0.016536
30 0.013728
35 0.0117
40 0.010478
45 0.009048
50 0.008112
55 0.007488
60 0.007098
65 0.00624
70 0.005928
75 0.005304
80 0.00507
85 0.00468
90 0.004368
95 0.004368
100 0.004056
};
\addlegendentry{1 gridpoints per cm}
\addplot [line width=1pt, green!92.681568627451!black, dashed, mark=*, mark size=2, mark options={solid}]
table {%
5 0.085332
10 0.041496
15 0.0273
20 0.021294
25 0.016536
30 0.013728
35 0.0117
40 0.010478
45 0.009048
50 0.008112
55 0.007488
60 0.007098
65 0.00624
70 0.005928
75 0.005304
80 0.00507
85 0.00468
90 0.004368
95 0.004368
100 0.004056
};
\addlegendentry{1.5 gridpoints per cm}
\addplot [line width=1pt, color3, dashed, mark=*, mark size=2, mark options={solid}]
table {%
5 0.087867123726
10 0.042640033326
15 0.028126862526
20 0.021393435637
25 0.016987160148
30 0.0141756552
35 0.012039181452
40 0.010582716326
45 0.00933905665200002
50 0.00843789
55 0.00765147865200002
60 0.00698195807400002
65 0.0064127964
70 0.0060752808
75 0.00562469747399999
80 0.00517820112599998
85 0.00495135385200002
90 0.0047252184
95 0.0043877028
100 0.00416654343700001
};
\addlegendentry{2 gridpoints per cm}
\addplot [line width=1pt, red!97.6476470588235!black, dashed, mark=*, mark size=2, mark options={solid}]
table {%
5 0.096174
10 0.04677075
15 0.029744
20 0.02197
25 0.018083
30 0.0149565
35 0.0127595
40 0.010985
45 0.009971
50 0.008957
55 0.008112
60 0.007267
65 0.0068445
70 0.006422
75 0.005915
80 0.005408
85 0.005239
90 0.004901
95 0.004732
100 0.004394
};
\addlegendentry{3 gridpoints per cm}
\addplot [line width=1pt, white!80!black, dashed, mark=*, mark size=2, mark options={solid}]
table {%
4.820540973 0.103190942
9.945453759 0.048036577
14.96357815 0.031594342
20.03675927 0.023458749
24.88853853 0.019161242
29.89834518 0.016063807
34.95543272 0.013617944
40.00209044 0.011923644
44.86185225 0.010874463
49.90457518 0.009809579
54.9275349 0.008740918
60.0166316 0.007982629
64.81548543 0.007325199
69.94159198 0.006963017
74.93712985 0.006329506
79.98509917 0.006029619
84.8531308 0.005578901
89.88384923 0.005624361
94.93181855 0.005332888
99.95256613 0.005014515
};
\addlegendentry{Fooladgar et al.}
\end{axis}

\end{tikzpicture}
\caption{Volume of cuboid uncertainty region versus baseline distance at different gridpoint spacings.}
\label{fig_cuboid-gridpoints}
\end{figure}

\begin{figure*}[!t]
\centering
\subfloat[]{%
     \scalebox{0.75}{
\begin{tikzpicture}

\begin{groupplot}[group style={group size=3 by 1,horizontal sep=50pt}]

\nextgroupplot[
tick align=outside,
tick pos=left,
x grid style={white!69.0196078431373!black},
xlabel={X (m)},
xmin=-100, xmax=100,
xtick style={color=black},
y grid style={white!69.0196078431373!black},
ylabel={Y (m)},
ymin=-100, ymax=100,
ytick style={color=black},
xticklabel style={/pgf/number format/.cd,fixed zerofill,precision=0},
y label style={at={(axis description cs:-0.1,.5)},rotate=0,anchor=south},
yticklabel style={/pgf/number format/.cd,fixed zerofill,precision=0},
axis equal image
]
\addplot graphics [xmin=-100, xmax=100, ymin=-100, ymax=100] {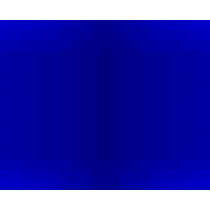};

\coordinate (top) at (rel axis cs:0,1);

\nextgroupplot[
tick align=outside,
tick pos=left,
x grid style={white!69.0196078431373!black},
xlabel={Z (m)},
xmin=0, xmax=200,
xtick style={color=black},
y grid style={white!69.0196078431373!black},
ylabel={X (m)},
ymin=-100, ymax=100,
ytick style={color=black},
xticklabel style={/pgf/number format/.cd,fixed zerofill,precision=0},
y label style={at={(axis description cs:-0.1,.5)},rotate=0,anchor=south},
yticklabel style={/pgf/number format/.cd,fixed zerofill,precision=0},
axis equal image
]
\addplot graphics [xmin=0, xmax=200, ymin=-100, ymax=100] {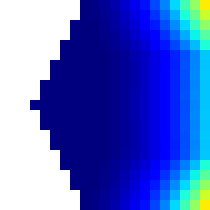};

\nextgroupplot[
tick align=outside,
tick pos=left,
x grid style={white!69.0196078431373!black},
xlabel={Z (m)},
xmin=0, xmax=200,
xtick style={color=black},
y grid style={white!69.0196078431373!black},
ylabel={Y (m)},
ymin=-100, ymax=100,
ytick style={color=black},
xticklabel style={/pgf/number format/.cd,fixed zerofill,precision=0},
y label style={at={(axis description cs:-0.1,.5)},rotate=0,anchor=south},
yticklabel style={/pgf/number format/.cd,fixed zerofill,precision=0},
axis equal image
]
\addplot graphics [xmin=0, xmax=200, ymin=-100, ymax=100] {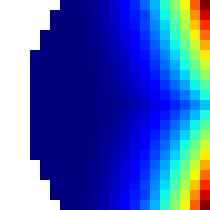};

\end{groupplot}

\coordinate (bot) at (25,2);


\begin{axis}[%
  hide axis,
  scale only axis,
  at={(bot)},
  anchor=east,
  xshift=0.25cm,
  point meta min=0.004565,
  point meta max=0.224155,
  colormap/jet,                     
  colorbar right,                  
  colorbar sampled,                     
  colorbar style={
  separate axis lines,
  samples=256,                        
  yticklabel style={
        /pgf/number format/fixed,
        /pgf/number format/precision=2
},
scaled y ticks=false
  },
  every colorbar/.append style={
    height=160
  },
  ylabel={Volume of Uncertainty Region (m\textsuperscript{3})}
]
  \addplot [draw=none] coordinates {(0,0)};
\end{axis}
\node [rotate=90] at (23,2.75) {Volume of Uncertainty Region (m\textsuperscript{3})};%
\end{tikzpicture}}  

\label{fig_box-planes}}
\vfil
\subfloat[]{%
     \scalebox{0.75}{
\begin{tikzpicture}

\begin{groupplot}[group style={group size=3 by 1,horizontal sep=50pt}]

\nextgroupplot[
tick align=outside,
tick pos=left,
x grid style={white!69.0196078431373!black},
xlabel={X (m)},
xmin=-100, xmax=100,
xtick style={color=black},
y grid style={white!69.0196078431373!black},
ylabel={Y (m)},
ymin=-100, ymax=100,
ytick style={color=black},
xticklabel style={/pgf/number format/.cd,fixed zerofill,precision=0},
y label style={at={(axis description cs:-0.1,.5)},rotate=0,anchor=south},
yticklabel style={/pgf/number format/.cd,fixed zerofill,precision=0},
axis equal image
]
\addplot graphics [xmin=-100, xmax=100, ymin=-100, ymax=100] {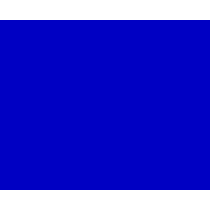};

\coordinate (top) at (rel axis cs:0,1);

\nextgroupplot[
tick align=outside,
tick pos=left,
x grid style={white!69.0196078431373!black},
xlabel={Z (m)},
xmin=0, xmax=200,
xtick style={color=black},
y grid style={white!69.0196078431373!black},
ylabel={X (m)},
ymin=-100, ymax=100,
ytick style={color=black},
xticklabel style={/pgf/number format/.cd,fixed zerofill,precision=0},
y label style={at={(axis description cs:-0.1,.5)},rotate=0,anchor=south},
yticklabel style={/pgf/number format/.cd,fixed zerofill,precision=0},
axis equal image
]
\addplot graphics [xmin=0, xmax=200, ymin=-100, ymax=100] {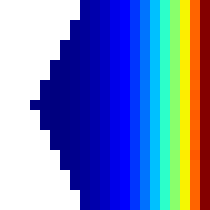};

\nextgroupplot[
tick align=outside,
tick pos=left,
x grid style={white!69.0196078431373!black},
xlabel={Z (m)},
xmin=0, xmax=200,
xtick style={color=black},
y grid style={white!69.0196078431373!black},
ylabel={Y (m)},
ymin=-100, ymax=100,
ytick style={color=black},
xticklabel style={/pgf/number format/.cd,fixed zerofill,precision=0},
y label style={at={(axis description cs:-0.1,.5)},rotate=0,anchor=south},
yticklabel style={/pgf/number format/.cd,fixed zerofill,precision=0},
axis equal image
]
\addplot graphics [xmin=0, xmax=200, ymin=-100, ymax=100] {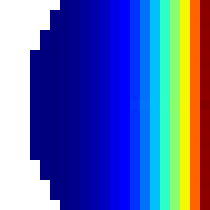};

\end{groupplot}

\coordinate (bot) at (25,2);


\begin{axis}[%
  hide axis,
  scale only axis,
  at={(bot)},
  anchor=east,
  xshift=0.25cm,
  point meta min=0.0000182,
  point meta max=0.0382,
  colormap/jet,                     
  colorbar right,                  
  colorbar sampled,                     
  colorbar style={
  separate axis lines,
  samples=256,                        
  yticklabel style={
        /pgf/number format/fixed,
        /pgf/number format/precision=2
},
scaled y ticks=false
  },
  every colorbar/.append style={
    height=160
  },
  ylabel={Volume of Uncertainty Region (m\textsuperscript{3})}
]
  \addplot [draw=none] coordinates {(0,0)};
\end{axis}
\node [rotate=90] at (23,2.75) {Volume of Uncertainty Region (m\textsuperscript{3})};%
\end{tikzpicture}}  
}
\label{fig_real-planes}
\caption{Volume of cuboid (a) and polyhedron (b) uncertainty region versus object positions within XY plane (left), XZ plane (middle), YZ plane (right)}
\end{figure*}
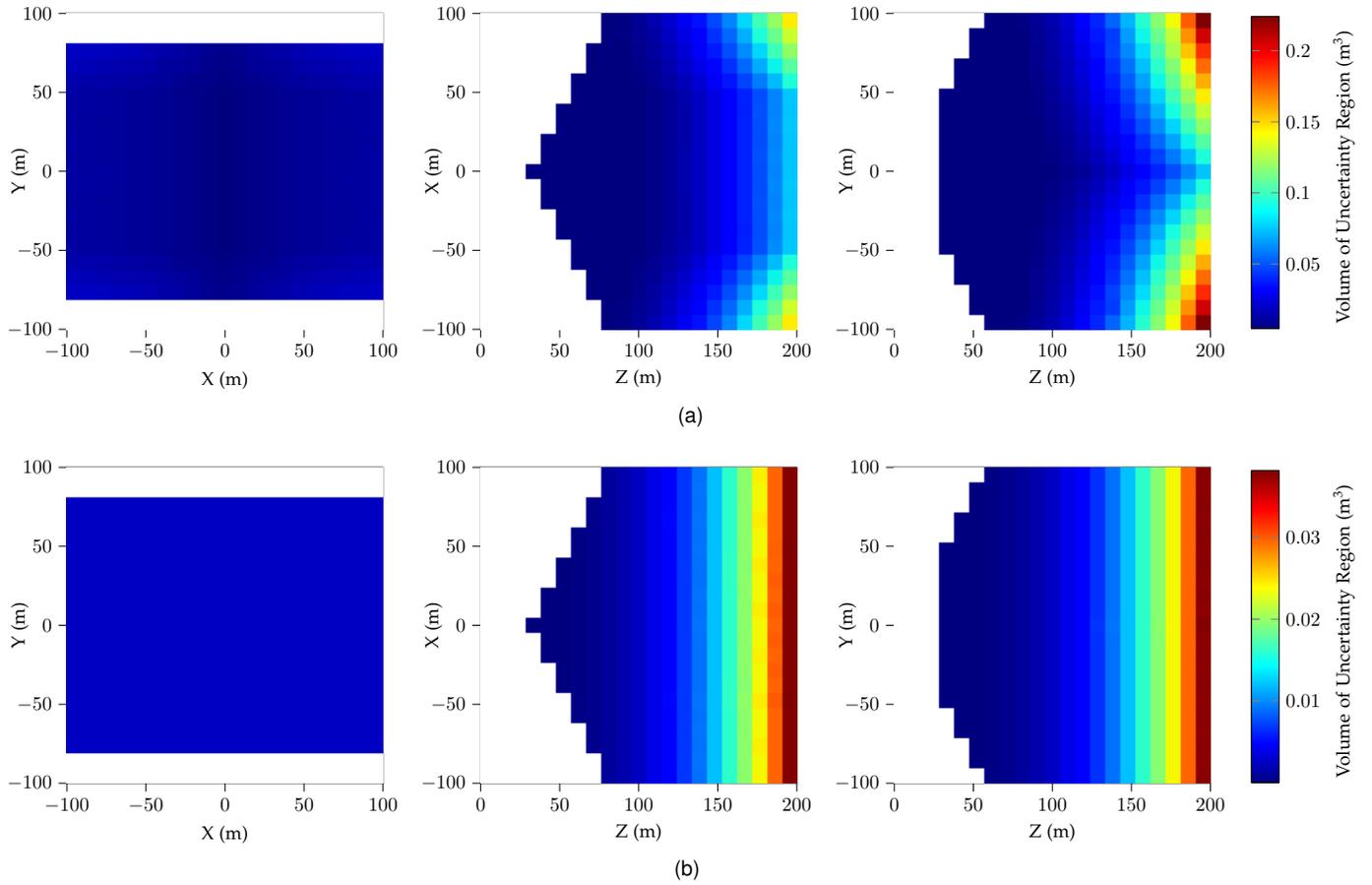

Fig. 9 shows the cuboid and polyhedron uncertainty volumes as a function of scene point position within three perpendicular planes that meet at (0, 0, 100).  
The first of the three plots in fig \ref{fig_box-planes} shows the same data as fig. \ref{fig_plane}, displayed as a 2-D image. The uncertainty volumes of points within the other two perpendicular planes of the same dimensions are also shown, colorized to the same scale. The variations in uncertainty volumes within the $XY$ planes in fig. \ref{fig_plane} are dwarfed by changes in the $XZ$ and $YZ$ planes over the same distances. Fig 9b displays the same plots using the polyhedron volume instead of the box volume approximation. The colorbar shows that the scale of all these volumes is much smaller. The variation in the $XY$ plane the previous authors reported is only present when using the cuboid volume approximation. For the $XZ$ plane, both volumes increase with $Z$, and the polyhedron volume is a factor of 4 smaller than the cuboid volume. For the $YZ$ plane both models show an increase in uncertainty volume with $Z$ but there is an added $y$ component in cuboid volume approximation that is not apparent in the polyhedron model.

\subsection{Grid Point Spacing}\label{sec:gridpoints}


The choice of how densely to generate the occupancy grid points strongly affects the calculated volume and the computation time. Here we perform the baseline experiment from the previous section with different grid point spacings ranging from 0.1 per cm to 3 per cm. Figure \ref{fig_cuboid-gridpoints} shows how as the number of scene points per cm is increased, the closer the results fit to the calculations by \cite{fooladgar_geometrical_2013} for the cuboid volume. The authors do not quantize their scene as we do in this method, so our cuboid volumes will never totally align but 3 points every cm at these baseline distances gives a visually close approximation and a total RMS error of uncertainty volumes of our cuboid model compared with the Fooladgar et al. model is 0.001808 $m^3$, shown in table 1. The comparison of our experiment results with theirs in terms of median symmetric accuracy shows that our inverse method produces similar results to their method. The same inverse method is used to calculate the polyhedron volume, so this suggests our polyhedron volumes are accurate. We use 3 grid points per cm for the experiments in this study unless stated otherwise.  

\section{Discussion}\label{sec:discussion}
The cuboid volume always overestimates the localization error when compared to the polygon volume, which is unsurprising given the visual representation in figures \ref{fig_pyramids} and 3. In all four experiments, we show this difference increases at larger uncertainty volumes. In all cases, the volumes increase when the size of the pyramids projecting from each pixel increases. This happens at larger pixel sizes, because this increases the pyramid cross section, and smaller focal lengths, which affects the distance between the pyramid apex and the pixels. The volumes also increase in size when the detecting pixel is closer to the center of the image plane. Even though all pixels are the same size and regularly distributed, perspective projection means that the outer pixels subtend smaller angles than those in the center. Larger baselines and smaller distances from the camera give lower localization error because in coplanar stereo cameras the scene point will be detected by pixels further away from the image center. 

\begin{figure*}[!t]
\centering
\subfloat[]{\includegraphics[width=3.5in, trim=90 50 40 60, clip]{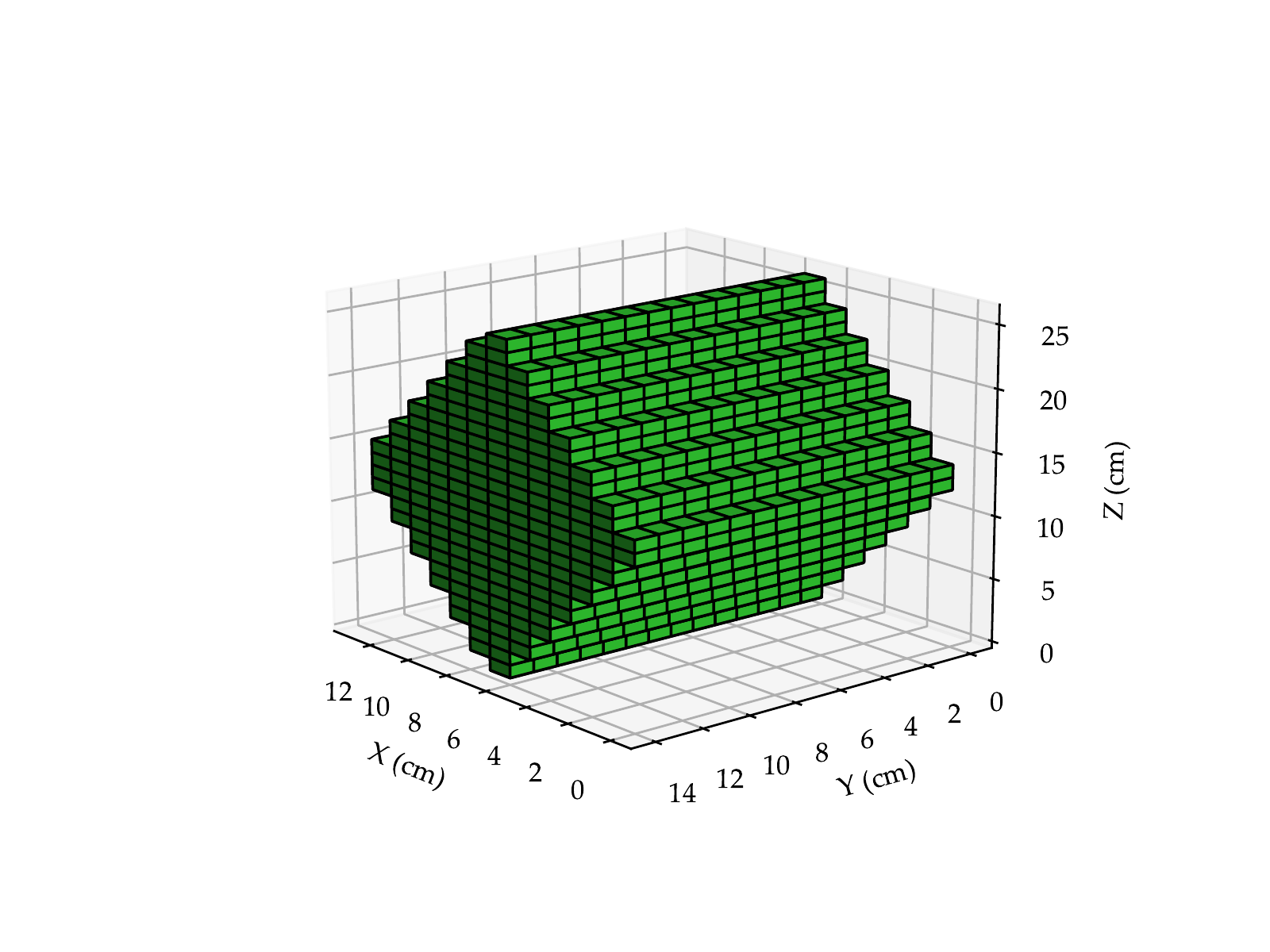}
\label{fig_voxels_1}}
\hfil
\subfloat[]{\includegraphics[width=3.5in, trim=90 50 40 60, clip]{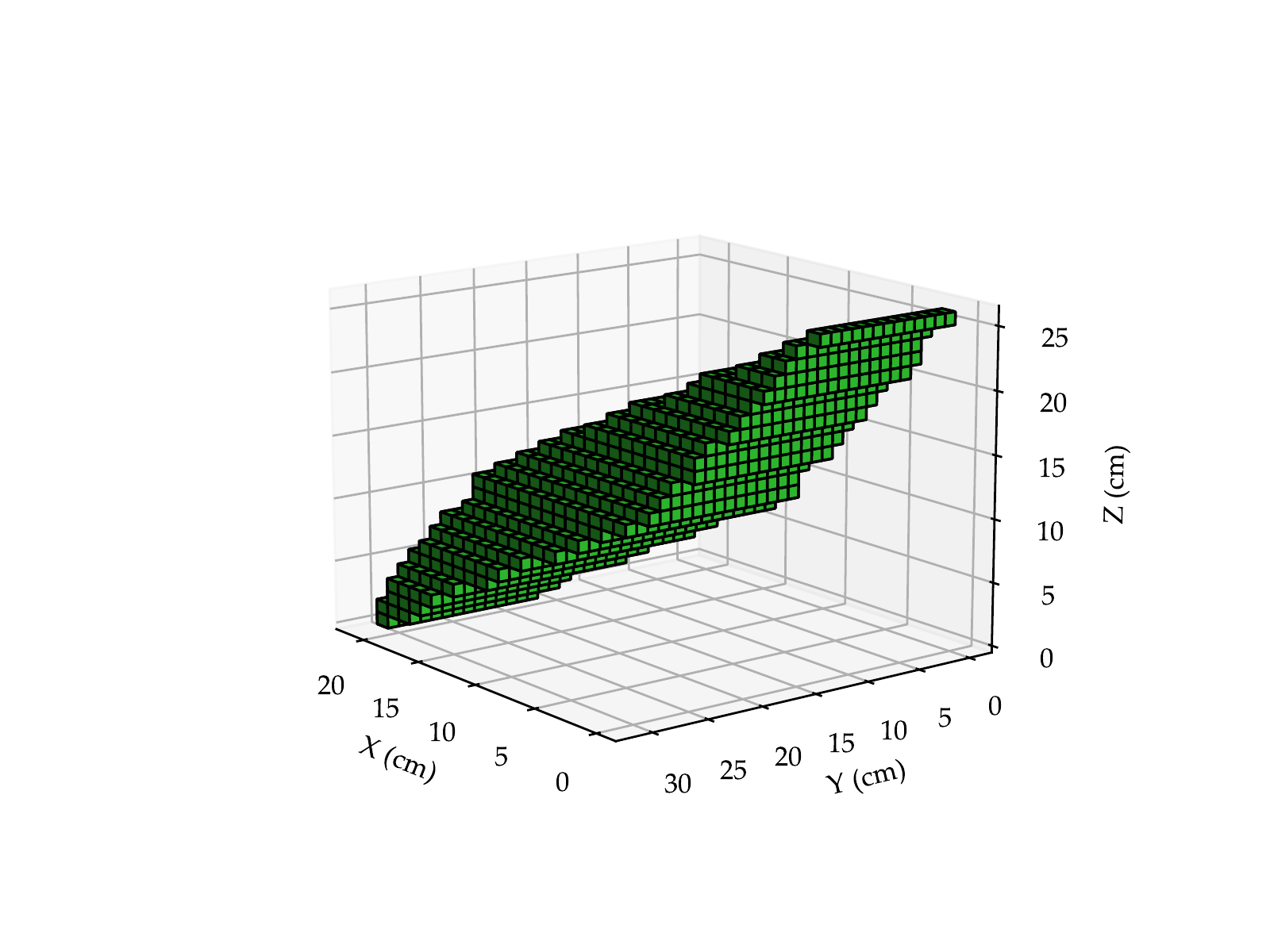}
\label{fig_voxels_2}}
\caption{Voxel plot of two intersections from fig. \ref{fig_plane}. Plot (a) shows an intersection from the center of the plane at $x=0$ m, $y=0$ m, Plot (b) shows an intersection from the corner at $x=-80$ m, $y=-80$ m. For visual clarity the gridpoints were generated at 1 per cm.}
\label{fig_voxels}
\end{figure*}

We might conclude that in a real-world scenario when taking stereo images of an object to make an accurate 3-D model, that the cameras should be aimed to detect the object at the edges rather than the center of the image plane, in order to minimize quantization uncertainty. However, we should also consider that lens distortion is at its greatest at the edges of the image, and at its minimum at the image center. Lens distortion is more widely considered as a cause of localization error and has been studied extensively in this field. This quantization error model could be combined with lens distortion models in the future to yield more comprehensive localization error predictions for a system. 

Previous authors reported variations in uncertainty volume as a function of scene point distance in $x$, however, after recreating these results with the cuboid volume model (figure \ref{fig_box-planes}), we show that there is no such trend in the polyhedron model (figure \ref{fig_real-planes}). We conclude that this trend is an artefact caused by the cuboid approximation. Our results show that only depth, $z$, affects the volume of the uncertainty region. A demonstration of the limitations of the cuboid model is shown in fig. \ref{fig_voxels}, which shows voxel plots of two example intersections from the XY plane in fig. \ref{fig_plane}. The uncertainty volume in fig. \ref{fig_voxels_1}, from the center of the $XY$ plane, contains 2548 gridpoints, which is converted to a polyhedron volume of 0.002548 $m^{3}$ and a cuboid volume of 0.0042 $m^{3}$ using (\ref{eq:volume}). The uncertainty volume in fig.\ref{fig_voxels_2}, from a corner of the $XY$ plane, contains 2362 gridpoints, equal to a polyhedron volume of 0.002362 $m^{3}$ and a cuboid volume of 0.015 $m^{3}$. The nature of the two shapes means that despite having similar polyhedron volumes their cuboid volumes are different by an order of magnitude. Intersections at the center of the $XY$ plane are more closely aligned with the coordinate system axes and the further away from the center the intersections are, the more oblique they become. This means that calculating the cuboid volume from the maxima and minima in gridpoint coordinates can yield a poor fit and result in significant overestimation of uncertainty due to quantization.

\section{Conclusion}


Quantization error is not usually factored into error predictions alongside errors from stereo matching and calibration. Wu et al. and Fooladgar et al. modelled quantization error by projecting pyramids or cones into space to calculate the intersection region for a pair of pixels, approximated as an ellipsoid or cuboid. They showed that quantization can introduce significant uncertainties in object feature mapping, however, under the computational restrictions of the time, it was only possible to calculate it for one pair of matched pixels. Our method applies their principle in reverse by quantizing the scene and calculating which pixels see each scene point. This method has the advantage of yielding every possible quantization region for every potential correspondence in the system, for two or more camera views. Furthermore rather than calculating a cuboid approximation of the uncertainty volumes, we calculate the volume of the uncertainty polyhedron itself. We confirmed earlier author's models, and show the cuboid approximation to overestimate the uncertainty volume by at least a factor of two. We present our method as a way to make quantization errors more readily calculable and more precise as they do not need to be approximated. 

The advantage of this calculation is that it only needs to be performed once to determine all possible quantization errors for a camera system. Then for example, at the correspondence matching stage of 3-D from stereo, the polyhedral uncertainty volume or cuboid volume dimensions can be queried and saved for each set of matched pixels. Furthermore the triangulation stage is no longer necessary because the object point locations are already known for each pixel correspondence. This has potential applications in mobile robot navigation, as most occupancy grid mapping methods that use camera inputs need to generate disparity maps. Using our method, a robot with multiple cameras would only need to store and query a look up table to determine scene geometry. Our method also calculates the visual hull on a pixel by pixel basis, so could also be used to construct visual hulls of larger objects spanning. A limitation of this method is that it assumes a static camera system, such as a stereo pair of cameras on a robot. As soon as the spacings of the cameras change, such as a camera moving through a scene, a new look up table has to be calculated. 

Another application of precisely calculated error bounds is in visualization. If we choose to represent the scene as a point cloud rather than an occupancy grid, we can plot the intersection polygons instead of points. Rather than needing to quote a positional error figure for each point, the localization uncertainty is implicit in the shape, making this easier for a human to interpret.


%

%
%
%
%
%
%

\ifCLASSOPTIONcaptionsoff
  \newpage
\fi



\bibliographystyle{IEEEtran}
\bibliography{References}
%
%
%

%

\newpage
\begin{IEEEbiography}[{\includegraphics[width=1in,height=1.25in,clip,keepaspectratio]{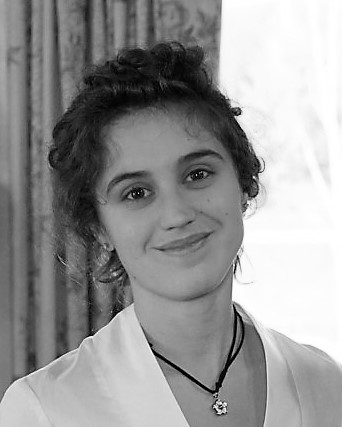}}]{Eleni Bohacek}

graduated with a first class degree in Earth Sciences from University College London (UCL) in 2014, with a specialism in Planetary Science. She then joined the UCL-Cambridge Centre for Doctoral Training in Integrated Photonic and Electronic Systems and completed her MRes in 2015. 
\par She is currently a PhD candidate at the Electronic and Electrical Engineering Department in collaboration with the Mullard Space Science Laboratory at UCL. Her thesis is focused on modelling uncertainty and errors in multi view stereo systems and building an emulator of the European Space Agency 2022 Rosalind Franklin rover camera systems.
\end{IEEEbiography}

\begin{IEEEbiography}
[{\includegraphics[width=1in,height=1.25in,clip,keepaspectratio]{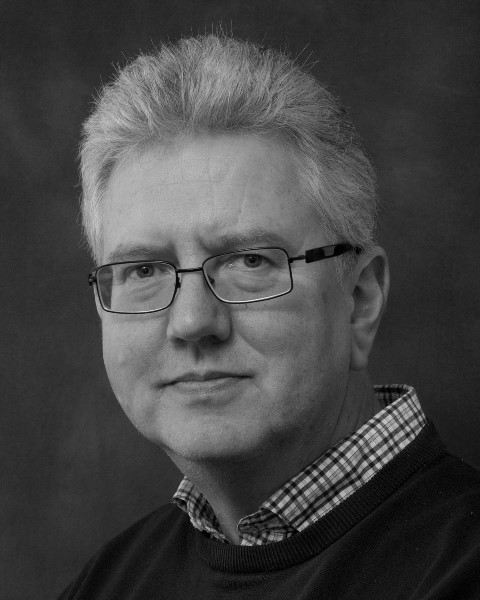}}]{Andrew J. Coates}
has been with Mullard Space Science Laboratory, University College London since 1982, with temporary positions at Max Planck Institute for Solar System Physics, Germany, University of Delaware, USA, and BBC World service (media fellowship). He is a Deputy Director (solar system) and Professor of Physics with UCL-MSSL. He is PI of the Rosalind Franklin (ExoMars 2022) PanCam team. He was a co-investigator on instruments for Cassini, Venus Express, Mars Express, and Giotto. He has authored/coauthored more than 525 papers with more than 425 refereed. He received the B.Sc. degree in physics from the University of Manchester Institute of Science and Technology, U.K., in 1978, and the M.Sc. degree and D.Phil. in plasma physics from Oxford University, U.K., in 1979 and 1982, respectively. He is a Fellow of the Royal Astronomical Society and member of the EGU and AGU. 
\end{IEEEbiography}

\begin{IEEEbiography}
[{\includegraphics[width=1in,height=1.25in,clip,keepaspectratio]{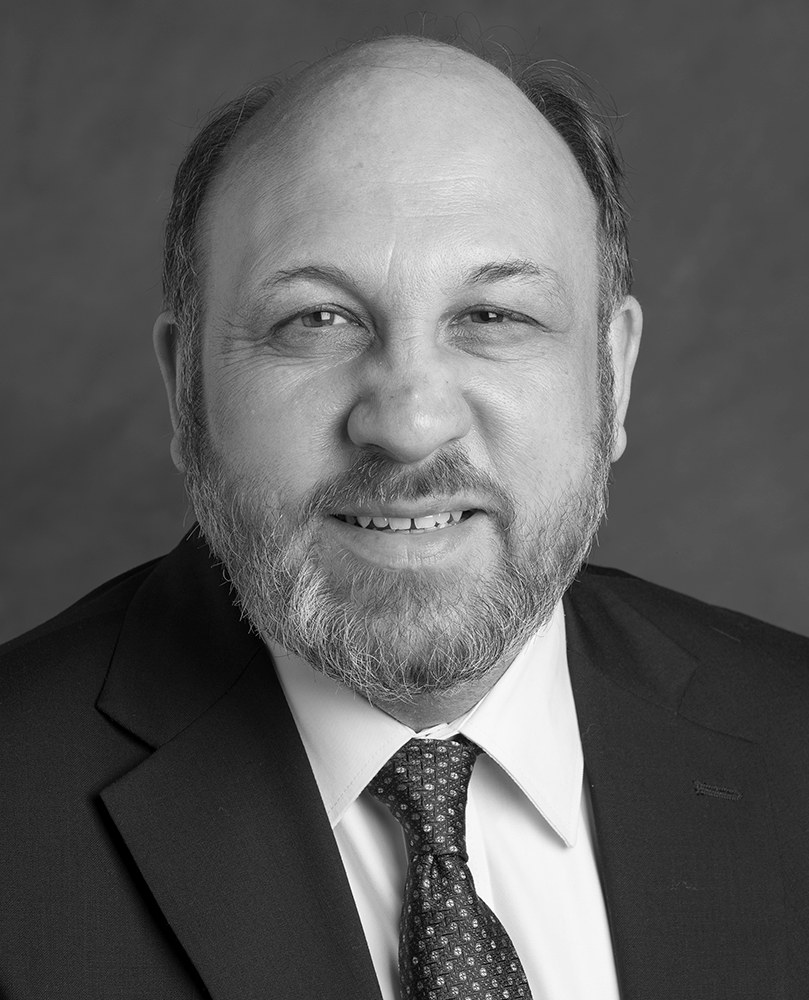}}]{David R. Selviah}
born in England, UK. He received the BA and MA degrees in physics and theoretical physics in 1980 and 1984 respectively and the PhD degree in photonic engineering in 2009 from Trinity College, Cambridge University, Cambridge. UK. He served internships at the Royal Aircraft Establishment, UK, Texas Instruments, UK and CERN, Switzerland. From 1980 to 1983, he was with the Allen Clark Research Center, Plessey (Caswell) Ltd., UK (now Oclaro). From 1983-1986, he was with the Department of Engineering Science, Oxford University, UK. He is currently a Reader in Optical Devices, Interconnects, Algorithms and Systems in the Optical Devices and Systems Laboratory of the Photonics Research Group in the Electronic and Electrical Engineering Department at University College London, UCL, London. He serves as a consultant to industry and is a founding director of the software company Correvate. He is the author of over 250 published articles, book chapters, keynote conference presentations and patents. His research is generally collaborative with international companies and universities. His current research interests include silicon photonic quantum dot lasers for high bit rate communication in data centers, signal processing, image processing, pattern and 3D object recognition, artificial intelligence, cloud computing, 3D Lidar and photogrammetry, 3D tracking, stimulation and monitoring neural behavior, data analysis for distributed acoustic sensors in oil and gas wells. Dr Selviah is a member of the Institute of Physics, Optical Society of America, European Optical Society and is a Chartered Physicist and Chartered Scientist. He represents the UK on the International Electrotechnical Commission Standards committees IEC TC86, SC86 WG4, SC86 WG6 and JWG9 covering optical fiber connectors, attenuation measurement techniques and optical circuit boards.
\end{IEEEbiography}

%




\end{document}